\documentclass[10pt,journal,final,twoside]{IEEEtran}
\usepackage{amsmath,amsfonts}
\usepackage{algorithmic}
\usepackage{algorithm}
\usepackage{array}
\usepackage[caption=false,font=normalsize,labelfont=sf,textfont=sf]{subfig}
\usepackage{textcomp}
\usepackage{stfloats}
\usepackage{url}
\usepackage{verbatim}
\usepackage{graphicx}
\usepackage{cite}
\usepackage{adjustbox}
\usepackage{lipsum}
\usepackage{color,soul}
\usepackage{xcolor}
\usepackage{nccmath}

\usepackage{hyperref}
\hypersetup{
    colorlinks=true,
    linkcolor=blue,
    citecolor=blue,
    urlcolor=blue
    }

\begin{document}

\title{SATA: Sparsity-Aware Training Accelerator \\ for Spiking Neural Networks}

\author{Ruokai Yin, 
Abhishek Moitra, \IEEEmembership{Graduate Student Member, IEEE, }\\Abhiroop Bhattacharjee, \IEEEmembership{Graduate Student Member, IEEE, } \\Youngeun Kim, \IEEEmembership{Graduate Student Member, IEEE, } \\and Priyadarshini Panda, \IEEEmembership{Member, IEEE} 
% \author{Ruokai Yin$^{\orcidlink{0000-0002-7550-0638}}$, 
% Abhishek Moitra$^{\orcidlink{0000-0002-0534-5206}}$, \IEEEmembership{Graduate Student Member, IEEE, }\\Abhiroop Bhattacharjee$^{\orcidlink{0000-0002-7721-271X}}$, \IEEEmembership{Graduate Student Member, IEEE, } \\Youngeun Kim$^{\orcidlink{0000-0002-3542-7720}}$, \IEEEmembership{Graduate Student Member, IEEE, } \\and Priyadarshini Panda$^{\orcidlink{0000-0002-4167-6782}}$, \IEEEmembership{Member, IEEE} 

\thanks{Manuscript received 13 March 2022; revised 29 May 2022 and 5 August 2022; accepted 14 September 2022. This work was supported in part by C-BRIC, one of six centers in JUMP, a Semiconductor Research Corporation (SRC) Program sponsored by DARPA; in part by the National Science Foundation under Grant 1947826; in part by DARPA AI Exploration (ShELL); and in part by Technology Innovation Institute (Abu Dhabi). This article was recommended by Associate Editor X. Lin. (\textit{Corresponding author: Ruokai Yin.})}
\thanks{The authors are with the Department of Electrical Engineering, Yale University, New Haven, CT 06511 USA (e-mail: ruokai.yin@yale.edu).}
\thanks{Digital Object Identifier 10.1109/TCAD.2022.3213211}
}


\maketitle

\begin{abstract}
Spiking Neural Networks (SNNs) have gained huge attention as a potential energy-efficient alternative to conventional Artificial Neural Networks (ANNs) due to their inherent high-sparsity activation. Recently, SNNs with backpropagation through time (BPTT) have achieved a higher accuracy result on image recognition tasks than other SNN training algorithms. Despite the success from the algorithm perspective, prior works neglect the evaluation of the hardware energy overheads of BPTT due to the lack of a hardware evaluation platform for this SNN training algorithm.
Moreover, although SNNs have long been seen as an energy-efficient counterpart of ANNs, a quantitative comparison between the training cost of SNNs and ANNs is missing. 
To address the aforementioned issues, in this work, we introduce SATA (Sparsity-Aware Training Accelerator), a BPTT-based training accelerator for SNNs.
The proposed SATA provides a simple and re-configurable systolic-based accelerator architecture, which makes it easy to analyze the training energy for BPTT-based SNN training algorithms.
By utilizing the sparsity, SATA increases its computation energy efficiency by $5.58 \times$ compared to the one without using sparsity.
Based on SATA, we show quantitative analyses of the energy efficiency of SNN training and compare the training cost of SNNs and ANNs.
The results show that, on Eyeriss-like systolic-based architecture, SNNs consume $1.27\times$ more total energy with considering sparsity (spikes, gradient of firing function, and gradient of membrane potential) when compared to ANNs.
We find that such high training energy cost is from time-repetitive convolution operations and data movements during backpropagation.
Moreover, to propel the future SNN training algorithm design, we 
%provide several observations on energy efficiency for different SNN-specific training parameters and 
propose an energy estimation framework for SNN training. \textit{Code for our framework is made publicly available.}
\end{abstract}


\section{Introduction}
\IEEEPARstart{R}{ecent} advances in deep learning have made artificial neural networks (ANNs) better candidates than humans for many tasks involving the processing of images, videos, and natural language \cite{schmidhuber2015deep}. Besides ANNs, Spiking neural networks (SNN), inspired by the processing paradigm of the human brain, are gaining popularity \cite{roy2019towards,schuman2022opportunities,christensen20222022}. SNNs primarily bring benefits to deep learning applications from two aspects: (1) the capture of both temporal and spatial information, whereas most ANNs lack the information from the time domain due to their spatial feedforward characteristics, (2) the energy-efficient implementations on hardware, since SNNs do not require multipliers for Multiply and Accumulate (MAC) operations during inference time. The inherent single-bit resolution of spikes also reduces the cost of memory communication.

Recently, there has been a growing interest in the field of SNN training algorithms. Works such as \cite{wu2018spatio,kim2020revisiting} have shown that Back Propagation Through Time (BPTT) \cite{werbos1990backpropagation} can achieve higher accuracy performance than spike-timing-dependent plasticity (STDP) \cite{stdp_article_1, stdp_artivle_2} and faster convergence than ANN-SNN conversion methods \cite{panda2020toward, roy2019towards, li2021free, li2022converting}. Despite the success from the algorithm perspective, these works neglect the evaluation of the hardware energy overheads of BPTT and thus fail to build the connection between the algorithm superiority and hardware efficiency in SNN training.

However, evaluating the hardware efficiency of SNN algorithms is not a direct task for algorithm researchers. Prior SNN algorithm works \cite{panda2020toward} use analytical methods to evaluate the hardware energy overheads of BPTT that neglect the underlying hardware architectural details leading to inaccurate estimations. In fact, a hardware evaluation platform for BPTT-based SNN training is missing in the SNN research community. Moreover, despite the fact that SNNs have been long treated as an energy-efficient counterpart of ANNs, there are very limited prior works comparing the energy difference between the two types of networks. 

In \cite{wang2020shenjing}, a rate encoding-based inference accelerator has been proposed and the inference energy for SNNs has been provided. However, the focus of the work is to optimize the NoCs for mapping SNNs onto the chip and has not given an energy comparison between SNNs and ANNs.
In the prior work \cite{narayanan2020spinalflow}, another inference accelerator for SNN has been proposed, however, the accelerator is based on temporal encoding which is different from the rate encoding that BPTT relies on. The work also provides the inference energy difference between SNNs and ANNs, however, a training energy comparison between two types of neural networks is still missing in the community. Recently, the work \cite{liang2021h2learn} has proposed a custom-tailored hardware architecture for SNN training that is highly SNN-tailored and targets performance boosting. For example, it utilizes LUT-based convolutions and has complex engines to compress the memory. With the complex and tailored design, it becomes difficult for researchers to make energy analyses of the different SNN training topologies on it. A fair comparison of the training energy cost between SNNs and ANNs is also hard to make on SNN-crafted architecture design. Hence, the work is unsuitable for general-purpose hardware evaluation of BPTT training. Moreover, they merely consider spike and spike gradient level sparsity that insufficiently captures the repercussions of BPTT on hardware.

\begin{table}[t]
\centering
    \caption{Comparison between SATA and other SNN accelerators work. $S$ denotes the spike activation, $\nabla f$ denotes the gradient of firing function and $\nabla U$ denotes the gradient of membrane potential.}
    \begin{adjustbox}{max width =\linewidth}
	\begin{tabular}{|l|c|c|c|}
        \hline
        \textbf{Accelerator}&\textbf{Type}&\textbf{Sparsity}&\textbf{Arch-design}\\
        \hline
        \hline
        Spinalflow \cite{narayanan2020spinalflow}&Inference&$S$&Systolic array-based\\
        \hline
        Shenjing \cite{wang2020shenjing}&Inference&$S$&SNN-crafted\\
        \hline
        H2Learn \cite{liang2021h2learn}&Training&$\nabla f$&SNN-crafted\\
        \hline
        SATA &Training&S, $\nabla f$, $\nabla U$ &Systolic array-based\\

        \hline
 	\end{tabular}\label{tb:related_works}
 	\end{adjustbox}
\end{table}

% \newpage

Motivated by the aforementioned problem, we propose SATA (Sparsity-Aware Training Accelerator), an Eyeriss-inspired \cite{chen2016eyeriss} general-purpose training accelerator for BPTT-based SNNs. The focus of SATA is to simulate a simple and re-configurable accelerator design, which simplifies the analysis of the training energy for BPTT-based training algorithms. Compared to prior works, SATA has several differences. Firstly, unlike prior works \cite{liang2021h2learn}, the SNN training architecture is more general and not overly optimized to a particular SNN architecture. This enables scalable hardware evaluation across a wide range of SNN models. Secondly, we show that sparsity in the gradients of membrane potential ($\nabla U$) can be leveraged to further improve the energy efficiency of SNN training. Moreover, Our general-purpose implementation approach additionally enables us to perform a fair comparison between ANN and SNN training. Finally, our training accelerator can be used as a benchmarking tool to evaluate the hardware training cost of SNNs. Table \ref{tb:related_works} summarizes our contributions with respect to prior digital SNN accelerator works that are most related to our works \cite{narayanan2020spinalflow,liang2021h2learn,wang2020shenjing}.

Another key point to optimize the energy efficiency of SNNs is to use the energy as a metric directly in training algorithm design. But today, a platform that can make a sparsity-aware estimation of the energy cost for SNN training is missing. We, therefore, propose a framework to estimate the computation and data movement energy in SNN training based on the architecture of SATA\footnote{Our framework is publicly available at \url{https://github.com/RuokaiYin/SATA_Sim}}. The framework extends the energy estimation method proposed in \cite{yang2017method} to further consider the impact of various groups of sparsity ($S$, $\nabla f$, and $\nabla U$) and SNN-specific training parameters, for example, the number of timesteps. 

We summarize our contributions as follows:
\begin{enumerate}

\item We present SATA, a sparsity-aware BPTT-based training accelerator for SNNs. The simple and highly re-configurable design makes it easy to perform a training energy analysis on SATA. The systolic array-based architecture also makes SATA the right baseline to make energy cost comparisons between SNN and ANN training. SATA also comprehensively captures three groups of sparsity (spike $S$, the gradient of firing function $\nabla f$, and the gradient of membrane potential $\nabla U$) to optimize the training energy efficiency. By utilizing those sparsities, SATA increases its computation energy efficiency by $5.58 \times$ compared to the one without using sparsity. Along with SATA, we also propose an energy estimation framework for SNN training based on SATA, which is publicly available \cite{satasim}.

\item We provide a training energy cost comparison between SNNs on SATA and ANNs on our baseline modified from the 8-bit version of Eyeriss\cite{chen2016eyeriss}. Our result shows that, on Eyriss-like systolic-based architecture, without considering sparsity for both SNNs and ANNs, SNNs consume $1.35\times$ more energy in total training energy compared to ANNs. Specifically, non-sparse SNNs consume $3.28\times$ more energy on computation and $1.28\times$ more energy on memory access compared to non-sparse ANNs. By further considering the sparsity ($S$, $\nabla f$, $\nabla U$), SNNs now consume $1.27\times$ more total training energy compared to ANNs. Specifically, sparse SNNs consume $1.19\times$ more energy on computation and $1.27\times$ more energy on memory access compared to sparse ANNs.

\item We also showcase various ablation studies on how the three groups of sparsity ($S$, $\nabla f$, $\nabla U$) change with different SNN training settings (for example, datasets, timestep, and network depth) and the training energy of SNNs resulting from the change of sparsity. We show that the total SNN training energy exponentially increases in a large timestep regime ($T > 32$). We also show that by having more sparsity in $\nabla U$, we can finally achieve less computation energy for SNN training compared to ANNs.

\end{enumerate}

\section{Related work}

There has been a wide range of works that have proposed accelerator designs to carry out SNN inference showing a high degree of parallelism, throughput, and energy-efficiency \cite{truenorth, davies2018loihi, ankit2017resparc, painkras2013spinnaker, kulkarni2019neuromorphic}. These include accelerators with a fully-digital architecture, such as IBM's TrueNorth processor \cite{truenorth}, as well as ones in which synaptic computational cores comprise of analog memristive crossbars, such as Resparc \cite{ankit2017resparc}. While most of the works focus on inference-only accelerator designs, some like Intel's Loihi processor account for SNN training using STDP learning rule \cite{roy2019towards, lee2018training}. Furthermore, the TrueNorth and Loihi processors are highly optimized to facilitate asynchronous spike communications with the objective of improving the performance of the deployed SNNs having a specific type of architecture, different from the conventional ones. However, they lack general applicability since they do not have support to benchmark a wide variety of SNNs, particularly SNNs trained by standard BPTT learning rules. Thus, it is imperative to have a general-purpose SNN training accelerator framework that can support the training and inference of a plethora of SNN architectures that is emerging from recent SNN algorithm studies. 

There is also a huge volume of work centered around SNNs that claim SNNs to be an energy-efficient alternative to ANNs due to high sparsity in input spikes \cite{panda2020toward, roy2019towards, ankit2017resparc, davies2018loihi, guo2019systolic}. But recently, an inference framework implemented in an \textit{Eyeriss}-like systolic-array hardware tailored for SNNs called SpinalFlow \cite{narayanan2020spinalflow} has shown that standard rate-coded SNNs with modest spike-rates exhibit significantly lower efficiency than corresponding accelerators for ANNs. Note, \textit{Eyeriss} \cite{chen2016eyeriss} follows a von-Neumann mode of neural computation widely adopted in modern accelerators and enables us to optimize over different design choices such as type of dataflow, computation reuse, and skipping zero computations. The primary cause behind the inefficiency of SNNs can be attributed to the storage and movement of membrane potentials over multiple timesteps during inference. With this in mind, the next steps include developing a similar hardware evaluation framework that can yield a realistic estimation of hardware energy and latency associated with training a wide range of SNN architectures over multiple timesteps. 

To this end, our SATA framework is the first to show that the inherent sparsity in SNNs associated with the spikes and their gradients are alone insufficient to yield training energy efficiency with respect to baseline ANN models. SNN training for conventional architectures, in fact, incurs huge overheads in terms of memory accesses and computations compared to ANNs, thereby making them highly energy-inefficient. Based on the conclusion and discussion posed in this work through the extensive study conducted on SATA and the energy-analysis tool that we propose, we hope that the future SNN algorithm research can be directed towards enhancing specific forms of sparsity (that impact computation cost largely) and avoiding certain values of SNN-specific training parameters (that impact memory cost largely) during training that can enable SNNs to be energy-efficient.

\section{Background}

\subsection{SNN Basics}
The network architectures for SNNs are very similar to that of ANNs, except that all ReLU-based neurons are replaced by simple neuron models to emulate biological neuron behaviors. This includes the update of membrane potential and the firing of spikes. Each pixel of input image fed to SNNs is encoded into a spike train that extends across the total timesteps $T$. Poisson distribution-based rate encoding scheme\cite{ahmed2016probabilistic} is primarily used \cite{kim2020revisiting,wu2018spatio}, where each pixel fires a spike train with a frequency proportional to its intensity. Noted that throughout the text, we refer to a timestep to the minimum time unit in SNN, in which a neuron updates the membrane potential according to the input and produces a spike if the threshold is reached.

\begin{figure}[t]

  \centering
  \includegraphics[width=0.9\linewidth]{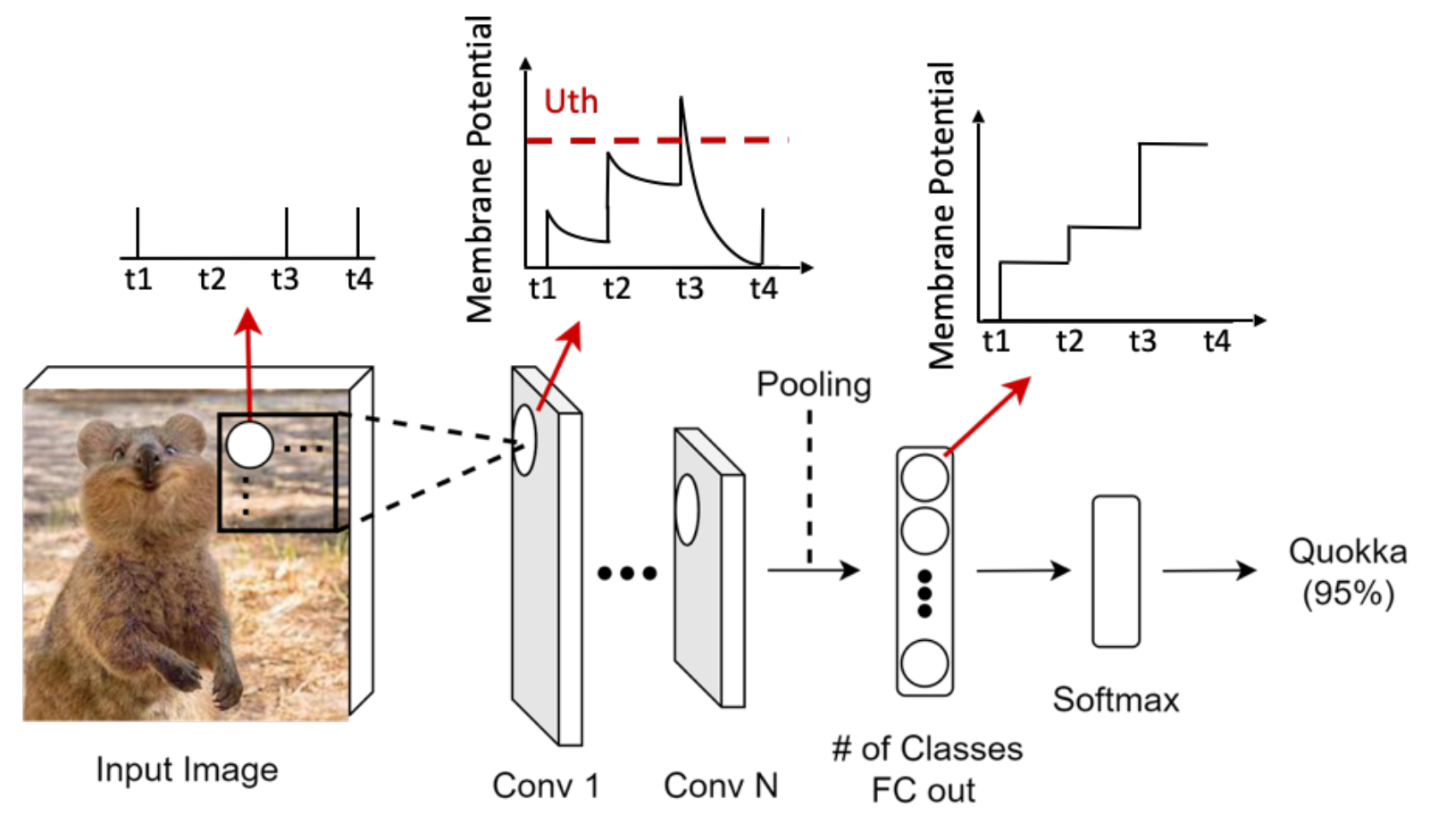}
  \vspace*{-3mm}
  \caption{An illustrative example of SNN. All intermediate neurons output rate-coded spikes and leak potential at each timestep $t$. Output neurons at the last fully connected layer will not generate spikes and only accumulate potential without leaking. }
  \label{fig:1}
\end{figure}

One of the most popular neuron models is a Leaky-Integrate-and-Fire (LIF) model. The LIF neuron receives binary spike inputs at every timestep $t$. After receiving the spikes, synaptic weights corresponding to each input spike are accumulated in the neuron's membrane potential $U$. The potential leaks at every timestep, based on the leaking factor $\alpha$. When the potential reaches the pre-set threshold $U_{th}$, the neuron fires an output spike and resets its membrane potential.
We model LIF using the explicit iterative expression: 
\begin{equation}
\label{eq:1}
U_{t}^{l}=
{\:}{\alpha}U_{t-1}^{l}(1-S_{t-1}^{l}) + W^{l-1}S_{t}^{l-1},
\end{equation}
\begin{equation}
\label{eq:2}
%   \sum_{i=0}^{\infty}x_i=\int_{0}^{\pi+2} f
    S_{t}^{l} = f(U_{t}^{l} - U_{th}),
\end{equation}
where $U_{t}^{l}$ and $S_{t}^{l}$ represent the potential and spike matrices of layer $l$ at timestep $t$. Also, $W^{l-1}$ is the weight matrix from previous layer $l-1$. And $f(\cdot)$ is the Heaviside step function, where $f(x) = 1$ when $x>0$, otherwise $f(x) = 0$. Fig. \ref{fig:1} shows an example SNN for an image classification task.

\begin{figure}[t]
  \centering
\includegraphics[width=0.45\linewidth]{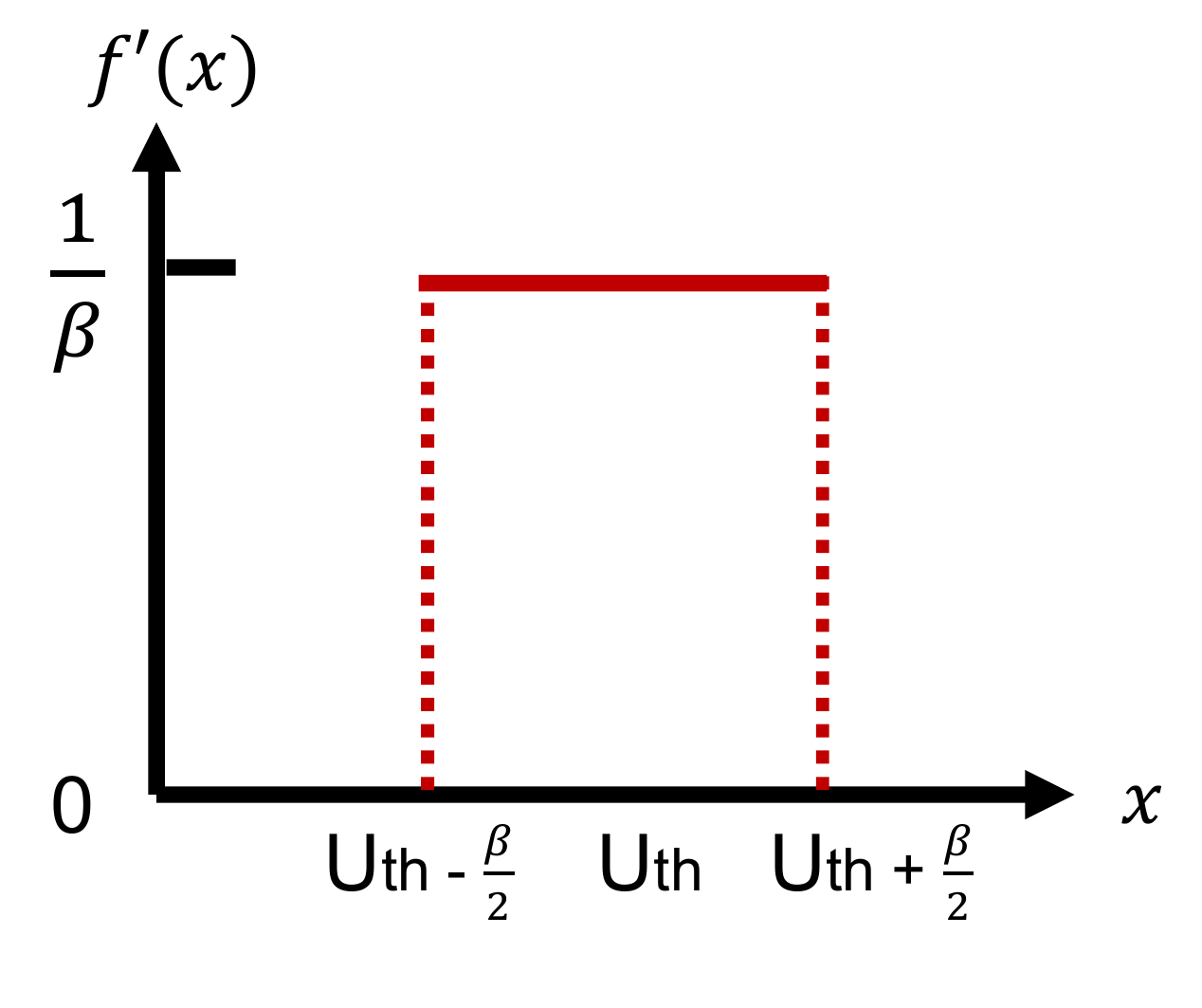}
\vspace{-5mm}
  \caption{Illustration of the curve to approximate the derivative of spike firing function. The derivative equals $\frac{1}{\beta}$ when the membrane potential is inside the $\beta$ range around the firing threshold $U_{th}$ during the forward propagation and equals zero otherwise. We name $\beta$ as the firing width.}
  \label{fig:firing_derivative}
\end{figure}

\begin{figure}[t]

  \centering
  \includegraphics[width=0.8\linewidth]{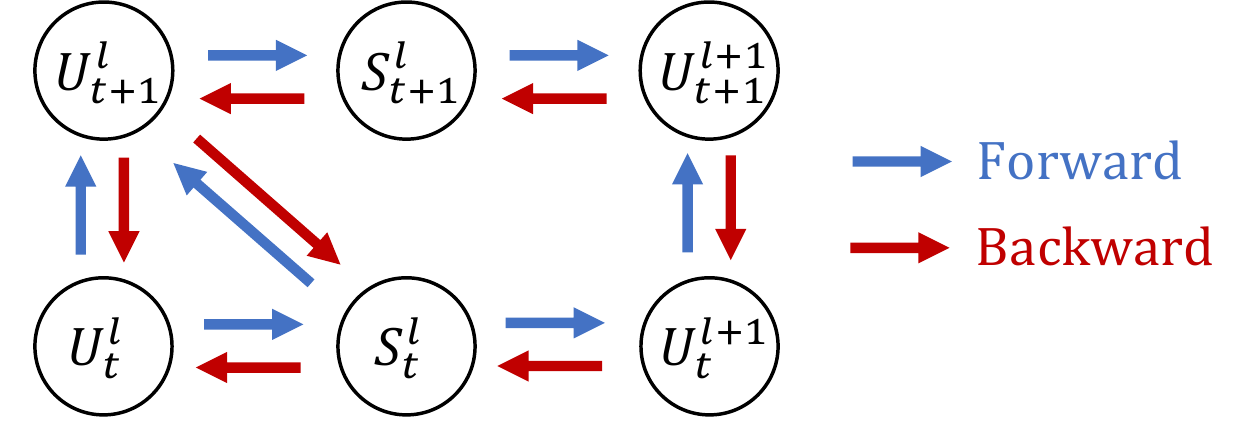}
  \caption{Illustration of how BPTT works. During the forward propagation, the neuron of layer $l$ at timestep $t+1$ will retain the potential $U_t^{l}$ and receive the spike $S_{t}^{l}$ from the previous timestep, which is considered as the propagation in the temporal domain. In the spatial domain, the neuron at layer $l+1$ receives the spike $S^{l}$ from the previous layer. Forward paths are shown in blue arrows. For the backpropagation, the paths are reversed and shown in red arrows.}
  \label{fig:2}
\end{figure}
% \vspace*{-5mm}

\subsection{BPTT for SNNs}

Recently, backpropagation through time (BPTT) algorithm\cite{werbos1990backpropagation, neftci2019surrogate} has become popular to train SNN models from scratch. BPTT shrinks the training accuracy gap between SNNs and ANNs by backpropagating gradients from both spatial and temporal domains, illustrated by Fig. \ref{fig:2}. The spike gradient  ${\nabla}S$ and potential gradient ${\nabla}U$ at layer $l$ and time $t$ with respect to loss function $L$ are expressed as:
\begin{equation}
\label{eq:3}
    {\nabla S}^{l}_{t}={\nabla U}^{l}_{t+1}(-{\alpha}U_{t}^{l})+{\nabla H}^{l+1}_{t},
\end{equation}
\begin{equation}
\label{eq:4}
    {\nabla U}^{l}_{t}={\nabla U}^{l}_{t+1}{\alpha}(1-S_{t}^{l})+{\nabla S}^{l}_{t}{f^{\prime}(U^{l}_{t})},
\end{equation}
 where ${\nabla H}^{l+1}_{t}$ represents the gradients backpropagated from the layer $l+1$ at timestep $t$, which can be formulated as:
\begin{equation}
\label{eq:44}
      {\nabla H}^{l+1}_{t}=W^{l+1}{\nabla U}^{l+1}_{t}.
\end{equation}
We use the function proposed in \cite{wu2018spatio} to approximate the derivative of Heaviside step function, where $f^{\prime}(x) = \frac{1}{\beta}$ when $|{x-U_{th}}|<\frac{\beta}{2}$, otherwise $f^{\prime}(x) = 0$. The approximated derivative of step function is illustrated in Fig. \ref{fig:firing_derivative}. The weight update for layer $l$ with learning rate $\gamma$ follows the rule below:
\begin{equation}
\label{eq:5}
    {W}^{l}=W^{l} - \gamma\sum_{t}{\nabla U}^{l}_{t}S_{t}^{l-1}.
\end{equation}

\section{Pitfalls and Opportunities in BPTT}
\label{sec:spa}
\subsection{Pitfalls in Memory Access and Computation}
Although the BPTT training algorithm boosts accuracy performance for SNNs, it deteriorates the hardware performance of the learning process by attaching memory consumption overheads. Since BPTT requires the information of spikes ($S$) and membrane potential ($U$) for every timestep during the forward propagation to conduct backpropagation, it introduces time-steps-related memory storage and communication overheads. As we will show in Section \ref{sec:exp}, the overhead scales exponentially with larger timesteps.

Besides memory overhead, energy overheads also exist in the SNN training computations. BPTT requires extra computations for updating the gradients ($\nabla H$) through layers by carrying out the same multi-bit multiply-accumulate (MAC) operation as ANNs and repeating it across all timesteps. The update of learnable parameters $W$ also repeats for each timestep to accumulate the temporal information. Besides the gradients of learnable parameters and activation, SNN also needs ancillary computations for gradients of membrane potential ($\nabla U$) as shown in Eqn. (\ref{eq:4}).

\subsection{Opportunities in Sparsity}

Fortunately, SNNs naturally exhibit high sparsity. By leveraging the sparsity in spikes $S$, we can reduce $\sim94\%$ of the MAC operations (reduced to accumulation-only operation in SNNs) in Eqn. (\ref{eq:1}) during the forward propagation of training (shown in Table \ref{tab: Sparsity_result} in Section \ref{sec:exp}). A similar number of gradients accumulated through timesteps in Eqn. (\ref{eq:5}) will also decrease.

As we discussed above, $f^{\prime}(U^{l}_{t}) = 0$ if $U^{l}_{t}$ is out of the $\beta$-width $U_{th}$ centered region. If $f^{\prime}(U^{l}_{t}) = 0$, we can skip Eqn. (\ref{eq:3}) that is the computation of $\nabla S_{t}^{l}$. Further, the add operation (corresponding to the second term in RHS) in Eqn. (\ref{eq:4}) as well as the fetch of $U_{t}^{l}$ can be eliminated. We define this as the sparsity in the gradient of the firing function ($\nabla f$).

Finally, if $\nabla U_{t+1}^{l} = 0$, we can skip the convolution computation of $\nabla H_{t}^{l}$. We define this as the sparsity in the gradient of potentials ($\nabla U$).
We summarize the sparsity-aware version of gradient calculation for membrane potential below:
\begin{equation}
\label{eq:6}
{\nabla U}^{l}_{t}=
\begin{cases}
{\:}{\alpha}{\nabla U}^{l}_{t+1}(1 - S_{t}^{l})
& \text{if sparsity in $\nabla f$}\\

{\nabla U}^{l}_{t+1}{\alpha}(1-S_{t}^{l})+{\nabla S}^{l}_{t}{f^{\prime}(U^{l}_{t})}
& \text{otherwise}
\end{cases}
\end{equation}
We will utilize these opportunities to guide the architecture design in the next section.

\begin{figure*}

    \centering
    \includegraphics[width=0.8\linewidth]{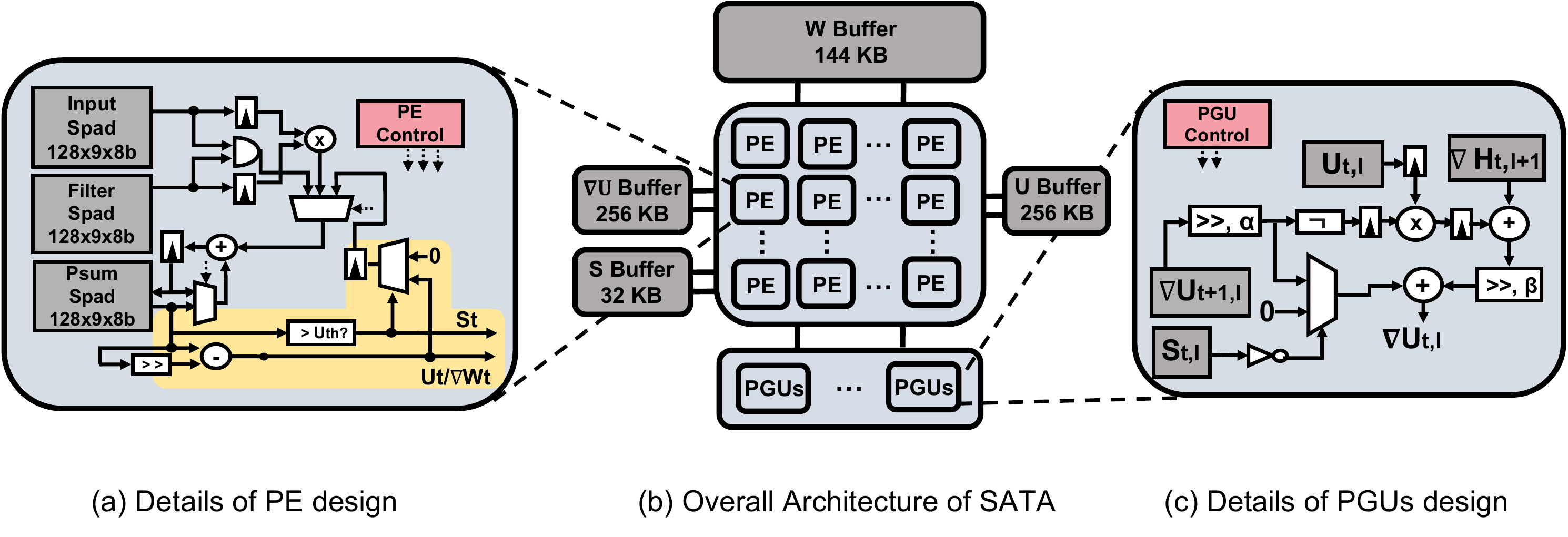}
    \vspace*{-5mm}
    \caption{Overview of SATA's architecture. (a) The dashed line indicates the signal for PE control. Each PE is equipped with a multiplier for the MAC operation during backpropagation and is attached to the circuit to carry out LIF computation during the forward computation (shaded in yellow). (b) The SATA architecture is composed of 128 PEs and 128 PGUs to facilitate the maximum number of output feature maps among any single layer in VGG5. And the different global buffers (GLBs) are also set to the corresponding size to facilitate maximum storage requirements among all layers in VGG5. (c) Each PGU composes of the circuits to carry out the computation of $\nabla U$ as in Eqn. (\ref{eq:3}, \ref{eq:4}).}
    \label{fig:4}
\end{figure*}

\section{Architecture Design}
\subsection{Architecture and Dataflow of SATA}

Similar to ANN training, convolution accounts for the majority of the computation workload in SNN training. Thus, we follow the spatial architecture design (doing MAC operations inside a processing element (PE) array) utilized by previous ANN accelerator works \cite{chen2016eyeriss,chen2014dadiannao} for SATA. However, the PE design for SATA needs to consider the difference in data representation and computation units across distinct convolution stages of SNN training, which will be explained later. Separate computation units for updating the gradients of membrane potential called potential gradient units (PGUs) are attached to simplify the design of PEs. Further, since computations in SNNs repeat for multiple timesteps, spatial dataflow that suits previous ANN accelerators, for example, row-stationary dataflow will no longer be energy efficient due to the repeated data communication cost between computation units and memory. To this end, SATA adopts a tailored temporal dataflow (namely, the combination of weight-stationary in \cite{chen2016eyeriss} and tick-batch in \cite{narayanan2020spinalflow}) for SNN-training to reduce the total energy overhead. We call this dataflow \textit{temporal weight stationary}.

In Fig. 5, we illustrate the temporal weight-stationary dataflow. The PE array has $K$ PEs and they first generate $T$ (total timesteps) outputs for all $K$ neurons that share the position (0,0) across $K$ output channels in the output feature map. Each PE only works on one output neuron. To maximally reuse the filters, each PE has a scratchpad to hold all the $C$ filters that participate in the computation at the corresponding output channel. First, $C$ input receptive fields (sized $R\times R$) for all the timesteps are fetched at once and shared by $K$ PEs. After the first computation cycle is done, all temporal and spatial computations required by those $K$ output neurons are completed. We will write them back to memory and fetch the next $C\times T$ input receptive fields to compute the $K\times T$ outputs for the next $K$ neurons. Notice that the same filters will stay in each PE and be reused until all the outputs are generated for the output feature map. By utilizing this dataflow, SATA fully reuses the filters across $T$ timesteps and reduces the repeated energy cost of each output neuron compared to non-temporal weight stationary dataflow.

\begin{figure}[h]

  \centering
  \includegraphics[width=0.8\linewidth]{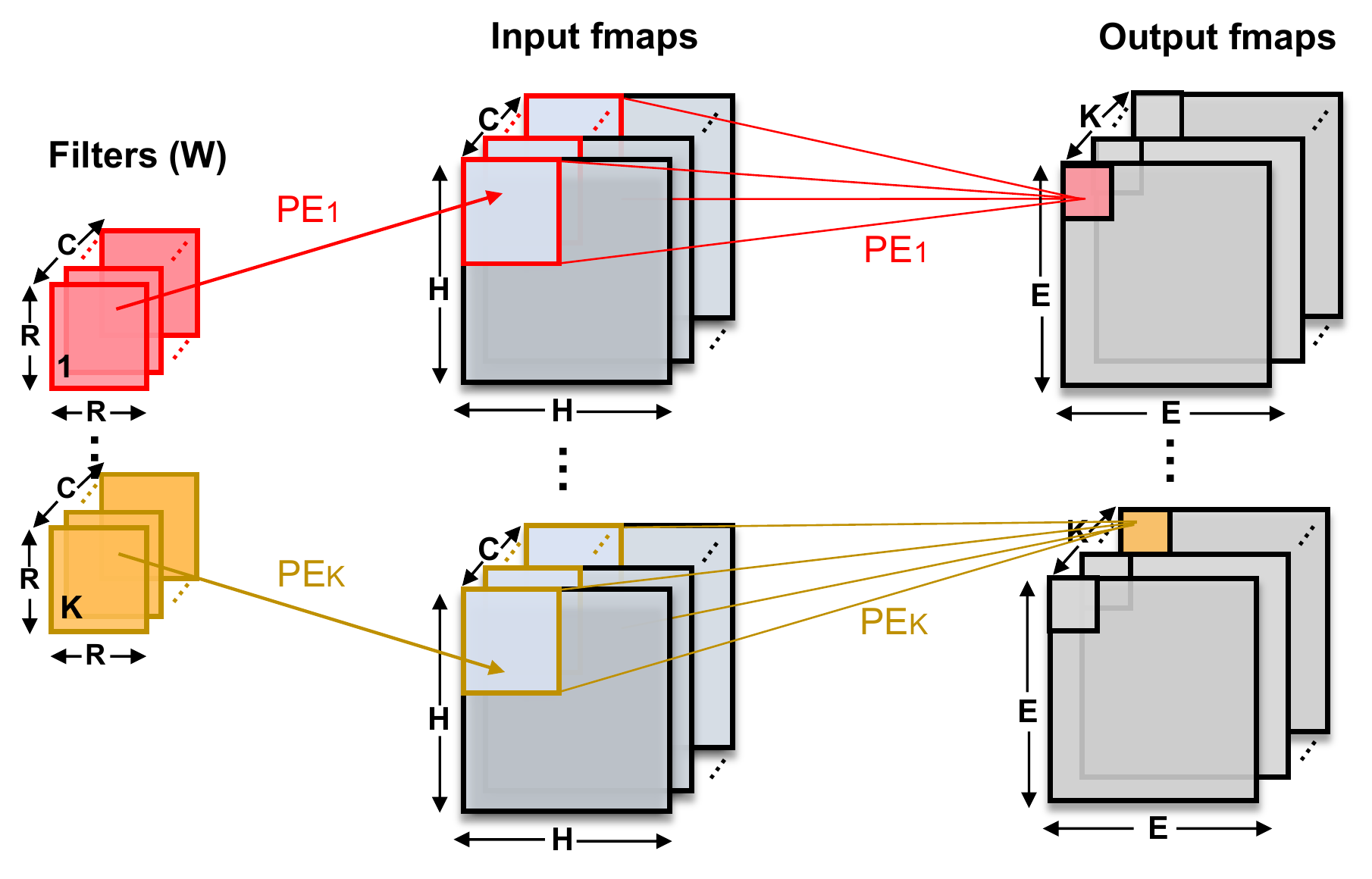}
  \vspace*{-3mm}
  \caption{Illustration of SATA's dataflow. The filters stay stationary in PEs for maximum filter reuse across timesteps. Further, each PE will only focus on the computation for one neuron in one output feature map at a time. For example, in the figure above, the pink-colored filters will be stored in PE$_1$ and PE$_1$ will be responsible for processing the pink-colored output pixel at the output feature map for all timesteps $T$.}
  \label{fig:5}
\end{figure}

The overall architecture for SATA is shown in Fig. 4(b). The example configuration considers training a VGG5 SNN with 8-bits resolution for all parameters in 8 timesteps \cite{kim2020revisiting}. We use 128 PEs and 128 PGUs in our design to facilitate the maximum number of feature maps in a single layer in VGG5. Generally, for other larger convolutional networks, the number of feature maps per layer is often a multiple of 128. We use a 144KB weight buffer to fit in the maximum number of 8-bit filters between two layers. $U$ and $\nabla U$ buffers are set to 256KB for holding 8-bit potentials and gradients for 128 neurons across all timesteps. Similarly, the $S$ buffer is set to 32KB due to the single-bit resolution of spikes.

\subsection{PE Design for Different Computation Stages}
There are three convolution stages in a complete cycle of SNN training: forward convolution, backpropagate convolution, and weight update convolution. The PEs are designed to be able to carry out the computations among all three stages as shown in Fig. 4(a). The filter scratch pads are set to the size of $128 \times 9 \times 8$ bits to be compatible with SATA's dataflow (considering most modern SNN architectures, like VGG, have $3\times3$ sized kernels). The other two scratch pads are set to the same size for making compatible computation with the filter's size.
\subsubsection{Forward Stage}
During the forward propagation, spike activations $S$ will be convolved with filters $W$ for all timesteps. Due to the 1-bit resolution of spikes, the multiplication will be simplified to and operations. At each timestep, after all the convolution partial sums are computed, the outputs go through the LIF computation units (yellow-shaded components in Fig. 4(a)) to generate spikes and update the membrane potential. If the input spike equals zero, the accumulation and the scratch pad read of filters will be elided.

\subsubsection{Backpropagation Stage}
To backpropagate gradients $\nabla H$ through the convolutional layers, convolutions are performed between 8-bit potential gradients $\nabla U$ and 8-bit filters $W$. Notice that during the backpropagation, $W$ needs to be transposed into $W^T$. This convolution is identical to the MAC-based convolution in ANN except for the repetition across all timesteps. Thus, we need an extra multiplier (see Fig. 4(a)) in the PE to accomplish the operation. The multiplier will be gated to save energy during the other two stages (namely, forward convolution and weight update stages). If the sparsity condition for $\nabla U$ is met, the MAC operation and scratch pad read of filters will be elided.

\subsubsection{Weight Update Stage}
Finally, spike activations $S$ stored during the forward propagation are convolved with potential gradients $\nabla U$ to generate the gradients for updating parameters $W$. This convolution reuses the computation units and the sparsity handling units from the forward propagation due to the identical data resolution. Again, this convolution needs to be repeated for all timesteps.

\subsection{Potential Gradient Units}
We use PGUs to accomplish the computation in Eqns. (\ref{eq:3}) and (\ref{eq:4}). The computation itself is straightforward, and we show the computation unit design in Fig. 4(c)). PGUs will first fetch $U_{t,l}$ to check whether there is sparsity in $\nabla f$. If the sparsity condition is satisfied, PGUs will omit the computation of $\nabla S$. Notice that one PGU will generate a single timestep $\nabla U$ for one neuron at a time. The number of PGUs can be configured to satisfy different throughput requirements. By default, SATA uses 128 PGUs.

\subsection{Discussion on Sparsity Handling}
\begin{figure}[t]
  \centering
  \includegraphics[width=0.9\linewidth]{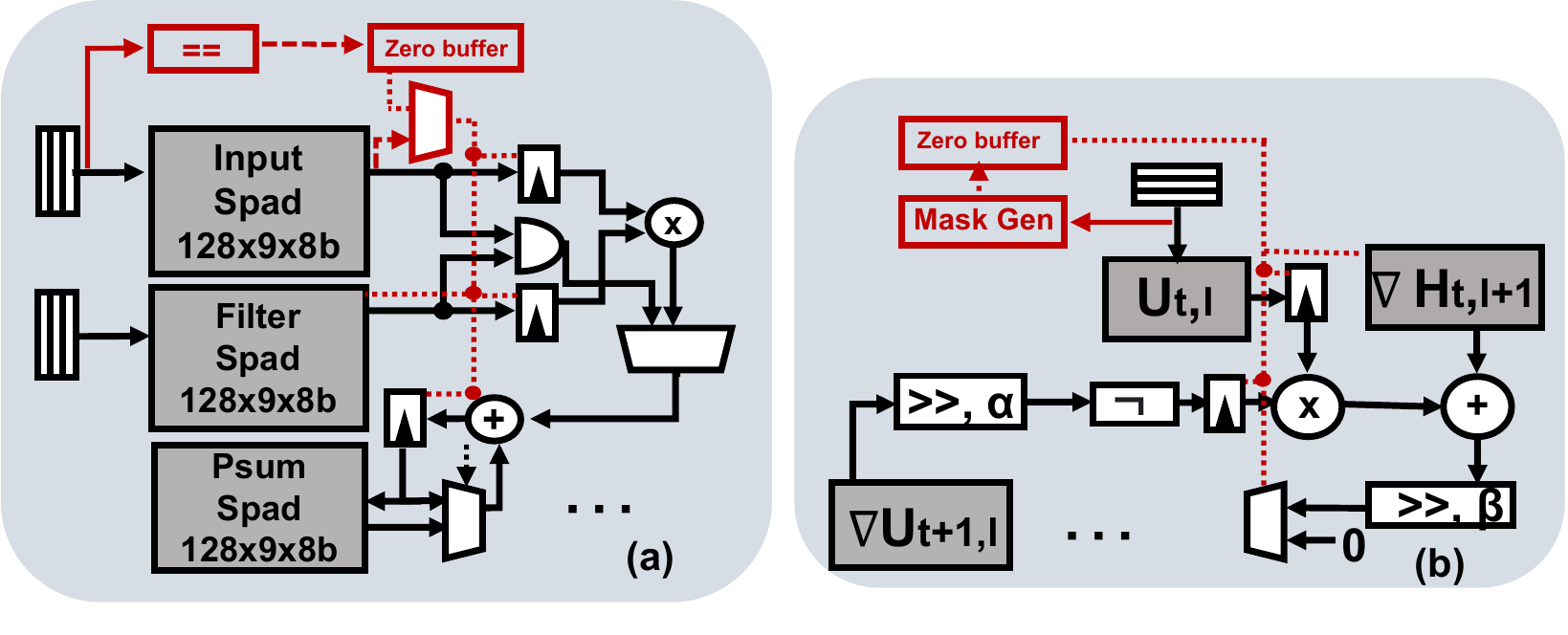}
  \vspace*{-3mm}
  \caption{Sparsity handling units inside PE and PGU. The red dash lines indicate the signal for sparsity handling units. We only show the units related to sparsity handling. (a) Leveraging input ($S$ and $\nabla U$) sparsity to save MAC and filter scratch pad reading. (b) Leveraging $\nabla f$ sparsity to skip the related computations of $\nabla S$.}
  \label{fig:sparsity_handle}
\end{figure}

In this subsection, we discuss the details of how we handle the sparsity inside PEs and PGUs and illustrate them in Fig. \ref{fig:sparsity_handle}. In general, we follow the gating method used in \cite{chen2016eyeriss_circuit} to omit the computation of MAC and memory read of filter scratch pad inside the PE when the input is zero. During the forward and weight update stage, we will directly use the spike input as the gating enable signal to disable the forward data path from switching and filter scratch pad from reading. During the backpropagation stage, similar gating logic will be applied, however, instead of directly using the input spike, we will use the bitmasks generated during the writing of the gradients to the input scratch pad. We have an extra 144-byte zero buffer to hold the bit masks.

In PGUs, we also apply a similar gating strategy as in PEs, however, this time we will check the $\nabla f$ sparsity condition as mentioned in section \ref{sec:spa}. During the writing of membrane potential $U$ into the scratchpad, the binary masks are generated by monitoring ${U_t}^l$, such that, mask $=1$ if $|{U_t}^l-U_{th}|< \beta/2$, else mask $=0$. Once the bitmasks are generated and stored in the zero buffer, we then use the same gating logic as in PEs to omit the multiplication and read of $\nabla H$ if the $\nabla f$ sparsity condition is met.

\begin{figure}[t]
  \centering
  \includegraphics[width=0.9\linewidth]{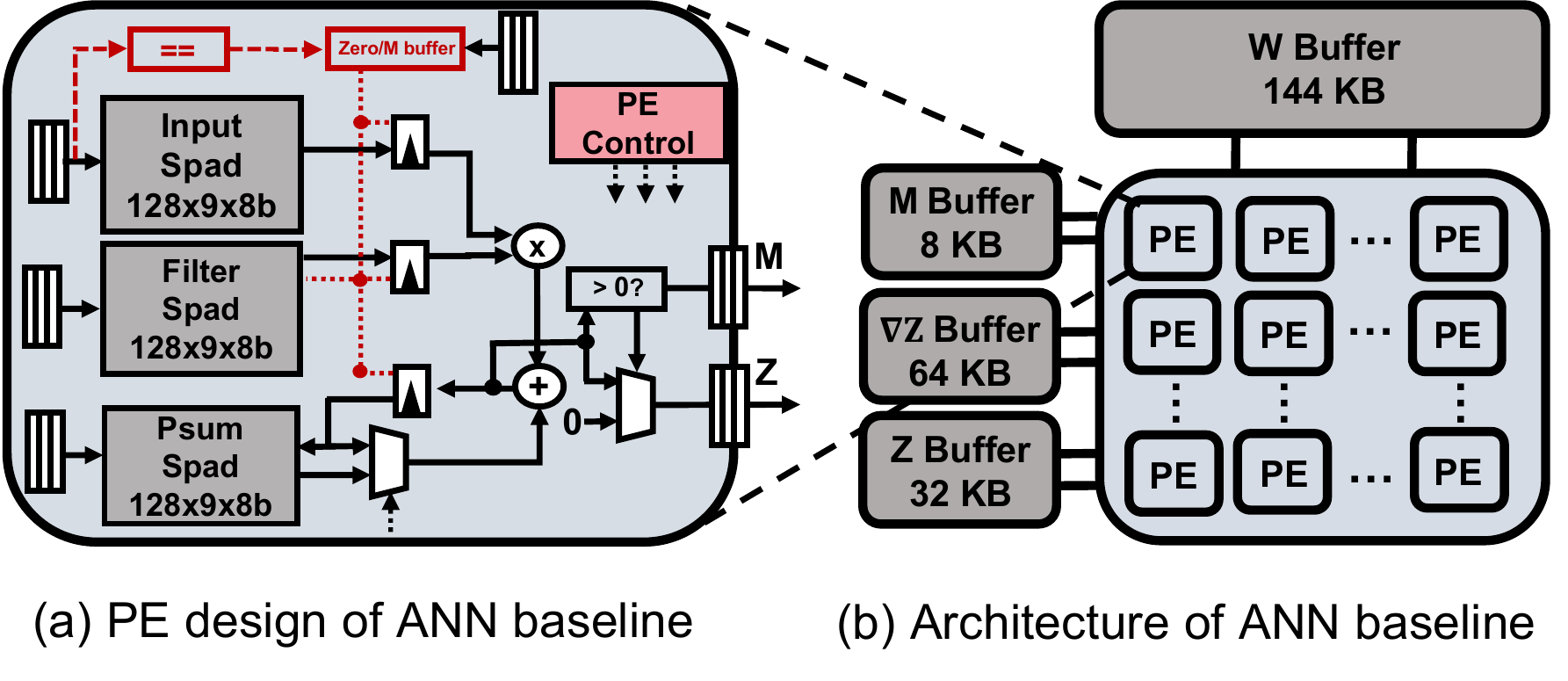}
  \vspace*{-3mm}
  \caption{Baseline architecture for ANN training. The PE and overall architecture design are based on Eyeriss, with additional hardware added to support backpropagation.}
  \label{fig:ann_baseline}
\end{figure}
\subsection{Architecture of ANN baseline}

To differentiate the training overhead of SNN (using BPTT algorithm) and ANN (standard backpropagation algorithm), we design a baseline architecture for standard backpropagation (BP)-based ANN training. The PE and architecture design is based on Eyeriss \cite{chen2016eyeriss}, an ANN inference accelerator that has the basic optimizations (reuse, zero-sliding, and memory hierarchy) that have been widely adopted in other ANN accelerator works \cite{goolgetpu}. In our baseline, we only attach necessary computation and memory components to the original design of Eyeriss to support BP-based training.

Inside the PE, we add a sign checker for carrying ReLU operation. The sign checker generates a bit-mask that is used during backpropagation to skip the unnecessary gradient computations (if the activation after ReLU is zero, we can skip the gradient calculation for that neuron during BP). Our baseline supports the same zero-sliding techniques as proposed in the original paper \cite{chen2016eyeriss}. A 64KB global buffer is added to hold the gradients during BP, together with an 8KB buffer to hold the masks that were generated during forwarding propagation. We illustrate the ANN baseline in Fig. \ref{fig:ann_baseline}

In the original Eyeriss paper \cite{chen2016eyeriss}, the \textit{Row-Stationary} dataflow is utilized to exploit spatial reuse of ifmaps, filters, and psums. However, a recent work \cite{narayanan2020spinalflow} has already shown that a rate-coded SNN is less energy efficient (up to $\sim60 \times$ more energy) when compared to the \textit{Row-Stationary}-based Eyeriss. As a result, we force our ANN baseline to use a similar dataflow to the one of SATA, which is more SNN friendly. Note, that SATA's dataflow does not bring any redundant memory or computation operations to the ANN baseline, which ensures a fair comparison.

\section{Energy Simulation Model}
\label{energy_simulation}
% \subsection{Model Overview}
In this section, we introduce our cost model for estimating the energy consumption of processing one single image based on SATA during SNN training. The total energy $E_{total}$ is the sum of three components: computation energy, memory energy, and the control circuit energy (noted as $E_{c}$, $E_{m}$, and $E_{ctrl}$). We further divide the computation energy into three stages as discussed above: forward computation energy, backward computation energy, and weight update computation energy ($E^{fwd}_{c}$, $E^{bwd}_{c}$, and $E^{wup}_{c}$). For the memory energy, we also divide it into three stages ($E^{fwd}_{m}$, $E^{bwd}_{m}$, and $E^{wup}_m$). The formula for total energy is shown below:

\begin{equation}
\begin{split}
    E_{total} &=(E_{c}^{fwd}+E_{c}^{bwd}+E_{c}^{wup})\\
    &+(E_{m}^{fwd}+E_{m}^{bwd}+E_{m}^{wup})+E_{ctrl}.\\
\end{split}\label{eq:7}
\end{equation}
We further divide the sub-stage energies into groups of sub-operation energy that belong to a given stage. More specifically, we divide the computation energy of the forward stage into the energy of MAC and LIF operation, the backward stage into the energy of MAC and $\nabla U$ calculation, and the weight update stage into the energy of MAC operation. For each calculation operation type, the energy of all the units along the computing path will be taken into consideration (for example, the energy will be different for the MAC operation in the backward stage and the other two stages, due to the different computation path). We also divide the memory energy of all three stages into the energy of communicating with DRAM, global buffers, and scratch pads.

The general rule for calculating those sub-operation energies is $N \times E$, where $N$ denotes the total number of the sub-stage operation that SATA requires to process one image and $E$ denotes the energy consumption of a single operation. Furthermore, we use $N(sp)$ to indicate that $N$ is the function of a given type of sparsity $sp$ (for example, $N_{mac}^{fwd}(sp_{S})$ is the total number of MAC operations during the forward propagation for SATA to process one image, and this number can be optimized by sparsity in $S$). We provide the energy cost estimation formula for all sub-stages as below:

\begin{equation}
\label{eq:8}
\begin{split}
    E_{c}^{fwd}&= N_{mac}^{fwd}(sp_{S}) \times E_{mac}^{fwd} + N_{LIF} \times E_{LIF},\\
    E_{c}^{bwd} &= N_{mac}^{bwd}(sp_{\nabla U}) \times E_{mac}^{bwd} \\ &+N_{\nabla U}(sp_{\nabla f})\times E_{\nabla U}, \\
    E_{c}^{wup} &= N_{mac}^{wup}(sp_{S}) \times E_{mac}^{wup}, \\
    E_{m}^{fwd} &= N_{dram}^{fwd} \times E_{dram} + N_{glb}^{fwd} \times E_{glb}\\
    &+ N_{spad}^{fwd}(sp_{S}) \times E_{spad}, \\
    E_{m}^{bwd} &= N_{dram}^{bwd} \times E_{dram} + N_{glb}^{bwd} \times E_{glb} \\
    &+ N_{spad}^{bwd}(sp_{\nabla f}) \times E_{spad},\\
    E_{m}^{wup} &= N_{dram}^{wup} \times E_{dram} + N_{glb}^{wup} \times E_{glb} \\
    &+ N_{spad}^{wup}(sp_{S}) \times E_{spad},
\end{split}
\end{equation}
where $E_{mac}^{fwd}$, $E_{mac}^{bwd}$, and $E_{mac}^{wup}$ denote the different energy of MAC operation in different sub-stage. $E_{LIF}$ denotes the energy of the LIF operation in the forward stage and $E_{\nabla U}$ denotes the energy of gradient calculation of $\nabla U$. $E_{dram}$, $E_{glb}$, and $E_{spad}$ denote the energy of a single time access to different memory units. The number of MAC operation in three stages are separately denoted as $N^{fwd}_{mac}$, $N^{bwd}_{mac}$, and $N^{wup}_{mac}$, which can be optimized by sparsity of $S$ and $\nabla U$. The number of LIF operations and calculation of $\nabla U$ (can be optimized by the sparsity of $\nabla f$) are also denoted by the corresponding $N$ notation. And the total number of data movement for three stages are denoted by the corresponding $N$ notation with the stage name on the top and the memory component name on the bottom, where the number of scratchpad reading can be optimized by $\nabla f$ and $S$. Note that we consider the data access of filters during the backpropagation into the weight update stage. 

In general, the number of computation operations is controlled by the network architecture of the SNN, while the number of memory movements will be determined by both the SNN network architecture and the hardware architecture and dataflow design. Table \ref{tab:sc_op_num} provides the total number of computation and data movement operations used in Eqn. \ref{eq:8} on SATA for VGG5 as an example and a reference.

\begin{table}[t]
\renewcommand*{\arraystretch}{1.6}
\caption{
The description of symbols used in Eqn. (\ref{eq:8}). The total number of each operation is calculated for a single image during one forward or backward propagation across all timesteps. Noted that we do not show the sparsity reduction of scratchpad accesses in the table for simplicity.}
\centering
    \begin{adjustbox}{max width =\linewidth}
	\begin{tabular}{|l|c|c|}
        \hline
        \textbf{Parameters}&\textbf{Description}\\
        \hline
        $C$&\# of input feature map or filter channels\\
        \hline
        $H$&input feature map width/height \\
        \hline
        $K$&\# of 3D filters or \# of output feature maps\\
        \hline
        $E$&output feature map width/height\\
        \hline
        $R$&filter width/height\\
        \hline
        $b$&maximum bitwidth (8 in SATA)\\
        \hline
        $T$&\# of timesteps (8 in SATA)\\
        \hline
        $sp_{S}$& spike sparsity (\# of zero spikes / \# of total spikes)\\
        \hline
        \vspace*{-1mm}
        $sp_{\nabla f}$&firing gradient sparsity\\
        &(\# of invalid spike grads / \# of total spike grads)\\
        \hline
        \vspace*{-1mm}
        $sp_{\nabla U}$&potential gradient sparsity\\
        &(\# of zero potential grads / \# of total potential grads)\\
        \hline
        \hline
        \textbf{\# of Ops}&\textbf{Description}\\
        \hline
        $N_{mac}^{fwd, wup}$& $T \times (1-sp_{S})\times \sum_{l=1}^{L}(C \times R^2 \times K \times E^2)_{l}$\\
        \hline
        $N_{LIF,\nabla U}$& $T \times \sum_{l=1}^{L}(K \times E^2)_{l}$\\
        \hline
        $N_{\nabla S}$& $T\times (1-sp_{\nabla f})\times \sum_{l=1}^{L}{(K \times E^2)_{l}}$\\
        \hline
        $N_{mac}^{bwd}$&$ T \times (1-sp_{\nabla U}) \times \sum_{l=1}^{L}(C \times R^2 \times K \times H^2)_{l}$\\
        \hline
        $N_{dram}^{fwd}$& $\sum_{l=1}^{L}(K \times C \times R^2)_{l}$\\
        & $+ T \times \sum_{l=1}^{L}(K \times E^2 + 1/b \times C \times H^2)_{l}$\\
        \hline
        $N_{glb}^{fwd}$& $2\times N_{dram}^{fwd}$\\
        \hline
        $N_{spad}^{fwd}$& $2\times\sum_{l=1}^{L}(K \times C \times R^2 + T \times 1/b\times C \times H^2)_l$\\
        \hline
        $N_{dram}^{bwd}$& $T\times \sum_{l=1}^{L}(K\times E^2 + 1/b \times C \times H^2)_{l}$\\
        \hline
        $N_{glb}^{bwd}$&$7\times T\times  \sum_{l=1}^{L}(K\times E^2)_{l}$\\
        & $ + \sum_{l=1}^{L}(2T\times 1/b \times C \times H^2 + K \times C \times R^2)_{l}$\\ 
        \hline
        $N_{spad}^{bwd}$&$\sum_{l=1}^{L}(K \times C \times R^2 + T\times K \times E^2)$\\
        \hline
        $N_{dram}^{wup}$&$2\times \sum_{l=1}^{L}(K \times C \times R^2)_{l}$\\
        \hline
        $N_{glb}^{wup}$&$2 \times (1+T)\times \sum_{l=1}^{L}(K \times C \times R^2)_{l}$\\
        &$ + T \times \sum_{l=1}^{L}( 1/b \times C \times H^2 + K \times E^2)_{l}$\\
        \hline
        $N_{spad}^{wup}$&$N_{glb}^{wup} + 2\times T \times \sum_{l=1}^{L}(K \times C \times R^2)$\\
        \hline
 	\end{tabular}\label{tab:sc_op_num}
 	\end{adjustbox}
\end{table}
\subsection{Energy model for considering the sparsity}
In Eqn. \ref{eq:8}, we define the total number of sparsity-related operations as a function of the sparsity. Then the user can define the abstraction level of the energy estimation results by setting the energy cost for a single operation $E$. For example, if one wants to test the theoretical maximum energy benefits that SATA can get from the sparsity, then $E$ can be set without considering any sparsity handling overhead. If the user wants to include the energy overheads of the sparsity handling units, it can be easily done by including the energy overheads into $E$. In Table \ref{tab:overhead_spa}, we give examples of the energy with and without sparsity handling units overheads. Then, the sparsity-aware energy with sparsity-handling overheads can be approximated by $N(sp) \times E(with overhead) + N \times E(overhead)$, where $E(overhead)$ can be calculated by simply subtracting the energy of operation without overheads from the one with overheads.
\begin{table}[t]
    
    \renewcommand*{\arraystretch}{1.5}
    \centering
    \caption{Energy difference for a single operation with and without overheads for sparsity handling units. The energy unit is normalized in terms of the energy for a MAC operation.}
    \begin{adjustbox}{max width =\linewidth}
	\begin{tabular}{|l|c|c|}
        \hline
        \textbf{Operation}&\textbf{Without Overhead}&\textbf{With Overhead}\\
        \hline
        \hline
        $E_{mac}^{fwd}$ &$0.146$&$0.146$\\
        \hline
        $E_{mac}^{bwd}$&$1.003$&$1.120$\\
        \hline
        $E_{\nabla U}$&$0.952$&$1.078$\\
        \hline

 	\end{tabular}\label{tab:overhead_spa}
 	\end{adjustbox}
\end{table}
\subsection{Discussion on Model Choice and Estimation Method}
In this section, we discuss our choice for the energy estimation model in Eqn. \ref{eq:7} and \ref{eq:8}. The goal of our energy model is to make it flexible and simple enough for users to adjust the complexity and accuracy of the energy model. For instance, as we will show in the later experiment setup, we choose to neglect the $E_{ctrl}$ in Eqn. \ref{eq:7} when we compare the training energy between SNNs and ANNs because the control energy would be approximately identical between SNNs and ANNs under the gradient-based training context. However, one can always apply the control energy to Eqn. \ref{eq:7} to make the energy value more accurate.

The estimation method used by our energy model is similar to the methodology proposed by \cite{yang2017method} and is verified in \cite{iccad_2019_accelergy}. Many prior works \cite{usystolic,kwon2019understanding,iccad_2019_accelergy,wu2020ugemm,wu2021ugemm} also follow this method to estimate the energy cost. Based on the prior works, we attach SNN-specific parameters (e.g., $T$ and $sp_{\nabla f}$) and consider SNN-specific operations (e.g., LIF and potential gradients update) to make the model work for SNNs. We can simply detach those efforts to make the model work for our ANN baseline.

\begin{table}[t]
\centering
    \caption{System parameters for SATA and Eyeriss, which are the baseline for SNNs and ANNs.}
    \begin{adjustbox}{max width =\linewidth}
	\begin{tabular}{|l|c|c|}
        \hline
        \textbf{Parameter}&\textbf{SATA}&\textbf{Eyeriss}\\
        \hline
        \hline
        Technology &65 nm CMOS&65 nm CMOS\\
        \hline
        Precision&8 bits (W, U), 1 bit (S, M)&8 bits (W, Z)\\
        \hline
        GLB size &256 KB (U, $\nabla U$)&32, 64 KB (Z, $\nabla$Z)\\
        &144 KB (W)&144 KB (W)\\
        &32 KB (S)& 8 KB (M)\\
        \hline
        Spad size&1.125 KB &1.125 KB\\
        \hline
        PE array size&128&128\\
        \hline
        PGU array size&128&-\\
        \hline
 	\end{tabular}\label{tab: Hardware_spec}
 	\end{adjustbox}
\end{table}

\section{Experiment results}
\label{sec:exp}
\subsection{Experiment Setup}
% The memory hierarchies configuration \cite{komuravelli2015stash,steinke2002assigning}
We use VGG5 \cite{vgg} (configured as in Table \ref{tab:network_config}) as our baseline network architecture for comparing the training energy difference between ANNs and SNNs. We train the VGG5 network on CIFAR10 with a learning rate of $0.001$, a momentum of $0.9$, and a weight decay factor of $1e^{-4}$. For SNN training, we further set the timestep as $T = 8$, leaking factor as $ \alpha = 0.94$ , firing threshold as $U_{th} = 0.75$, and the fire function width $\beta = 2.5$. 

We use SATA-Sim \cite{satasim} with the energy simulation model in \ref{energy_simulation} to approximate the training energy of ANNs from the 8-bit version of our Eyeriss-based ANN baseline and SNN from SATA both with the computing units synthesized in Synopsys Design Compiler at 400MHz using 65nm CMOS technology and the memory units simulated in CACTI\cite{muralimanohar2009cacti}. Since the main purpose of the energy results is for comparison, we assume perfect gating and no control overheads during the comparison (namely, assuming no leaking power for computation units when gated and setting $E_{ctrl}$ in Eqn. (\ref{eq:7}) to 0 for both ANN and SNN). Unless otherwise stated, the hardware specifications are listed in Table \ref{tab: Hardware_spec}. All the energy results denote the energy required to process one image and the unit of energy is normalized in terms of the energy for a MAC operation (e.g., $100$ = energy of $100$ MAC operations).
 
For performing energy analysis on sparse training, the inherent sparsity is collected for both SNN and ANN baseline during the training process. We collect the layerwise sparsity of activation (arising due to ReLU non-linearity which only passes non-negative values) and its gradient for ANNs and collect three categories of sparsity (namely, $S$, $\nabla f$, and $\nabla U$) for SNNs. All the SNN sparsity results are averaged across total timesteps, the number of images, and training epochs. The sparsity results are summarized in Table \ref{tab: Sparsity_result}.

\begin{table}[t]
\centering
    \caption{Network Structures for VGG5 and VGG9. The symbols C, MP, and FC denote convolutional, max-pooling, and fully connected layers, respectively. 64C3 refers to a convolutional layer with 64 channels and 3$\times$3 kernels.}
	\begin{tabular}{|l|c|c|}
        \hline \textbf{Network}&\textbf{Structure}&\textbf{Dataset}\\
        \hline
        \hline
        &64C3-MP2-128C3-128C3-&MNIST\\
        VGG5&MP2-1024FC-10FC&CIFAR10\\
        &&CIFAR100\\
        \hline
        &64C3-64C3-MP2-128C3-&\\
        VGG9&128C3-MP2-256C3-&CIFAR10\\
        &256C3-256C3-MP2-1024FC-10FC&\\
        \hline
 	\end{tabular}\label{tab:network_config}
\end{table}

\subsection{Training Energy: SNNs vs. ANNs}
\begin{figure}[t]
  \centering
  \includegraphics[width=\linewidth]{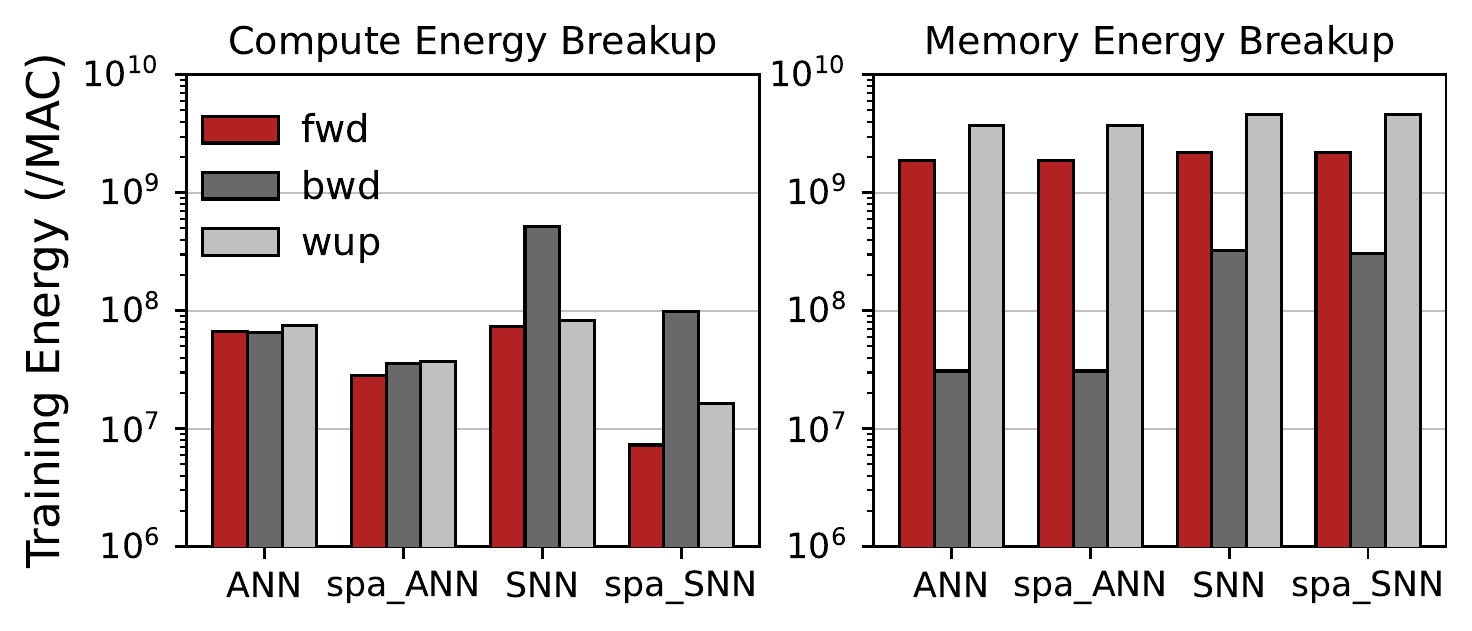}
  \vspace*{-7mm}
  \caption{Energy comparison between ANNs and SNNs, where spa-ANN and spa-SNN refer to sparse ANN and sparse SNN, respectively.}
  \label{fig:snn_ann_comp}
\end{figure}

We first compare the training energy between SNNs and ANNs without considering any sparsity in Fig. \ref{fig:snn_ann_comp}. In our training scenario, SNN in total consumes $1.35\times$ more energy than ANN.
We further break up the energy comparison results into computation energy and memory energy. According to our comparison, SNN consumes $3.28\times$ more total computation energy when compared to ANN and $1.28\times$ more total memory movement energy compared to ANN.

\begin{figure}[t]
  \centering
    \includegraphics[width=0.85\linewidth]{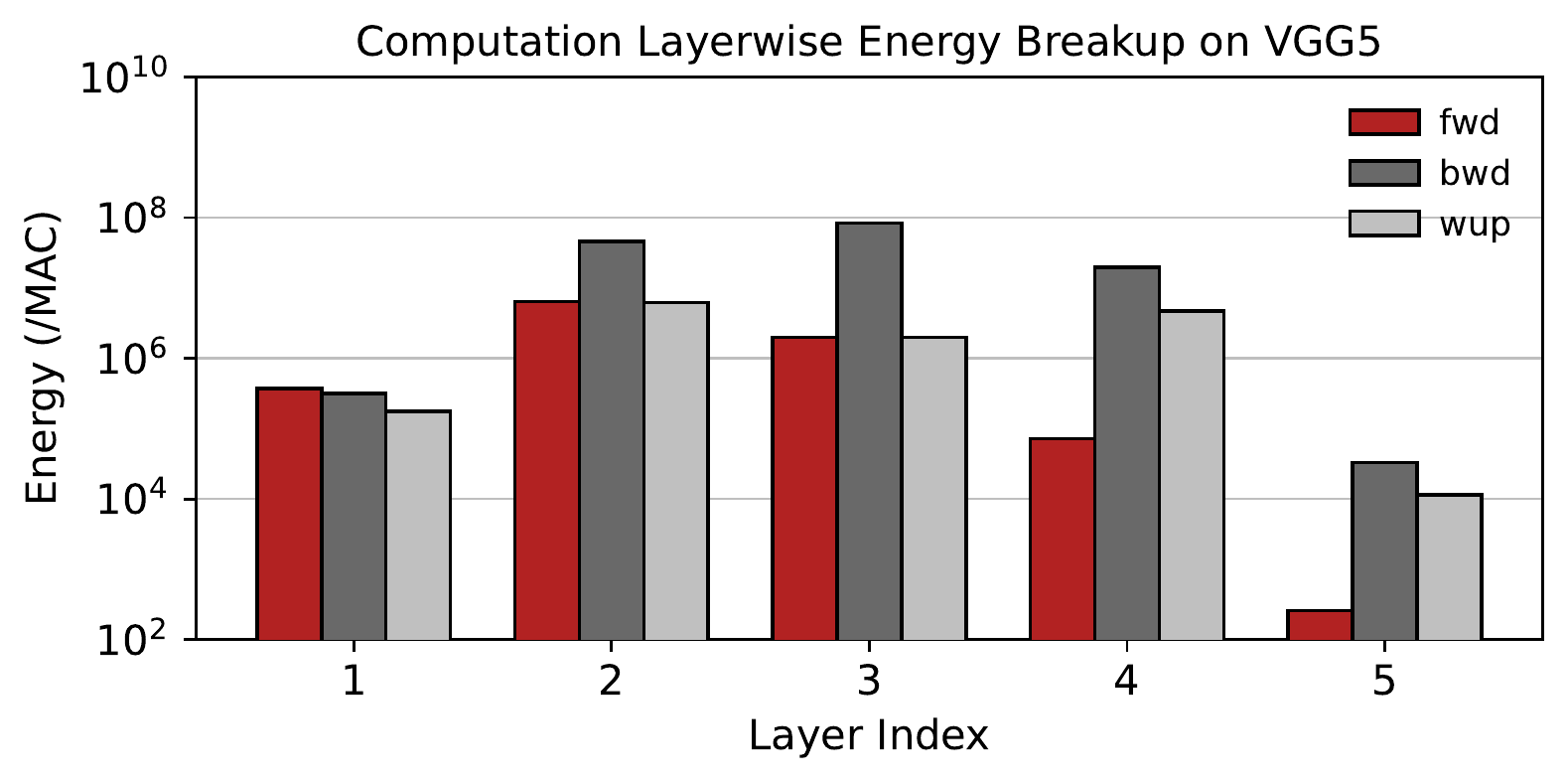}
\vspace*{-5mm}
  \caption{Layerwise computation energy results on VGG5.}
  \label{fig:layerwise_vgg5}
\end{figure}

\begin{figure}[t]
  \centering
  \includegraphics[width=\linewidth]{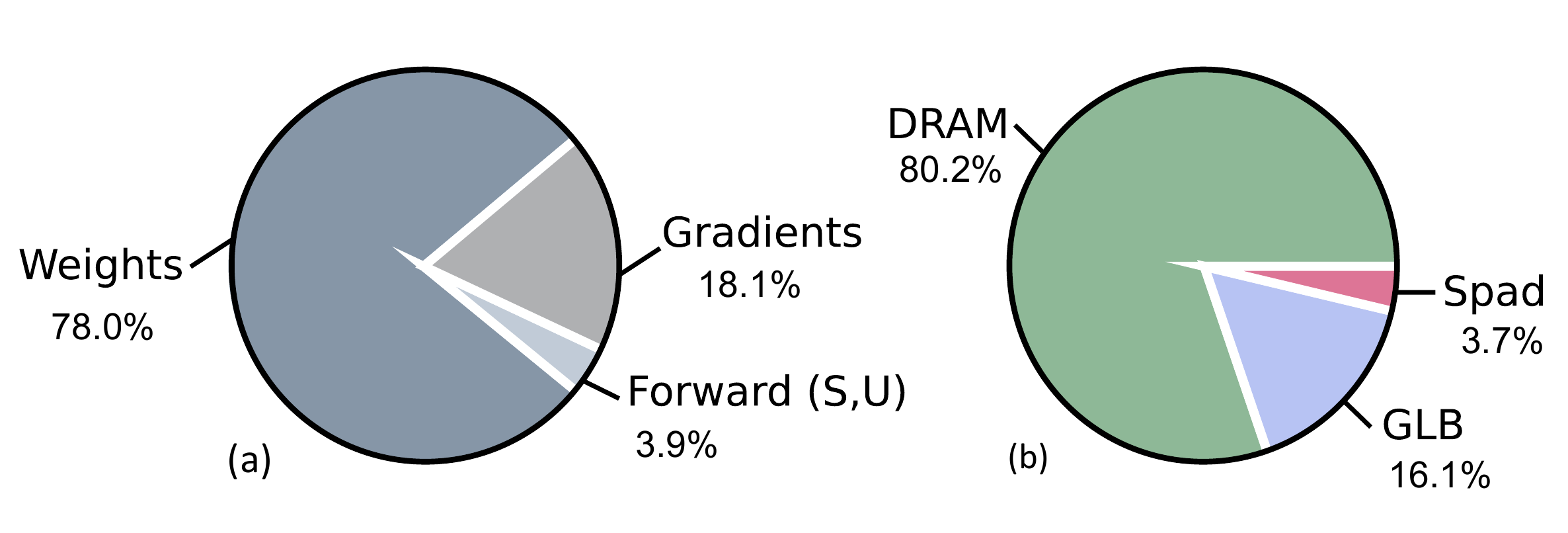}
  \vspace*{-6mm}
  \caption{Energy breakdown of the memory for VGG5 from the perspective of (a) algorithm memory components and (b) hardware memory components. In (a), the Gradients refer to the memory movement to calculate gradients ($\nabla U$, $\nabla H$, and $\nabla W$).}
  \label{fig:memory_break}
\end{figure}

We then take sparsity into consideration. The sparsity results can be found in Table \ref{tab: Sparsity_result} for SNNs and ANNs for CIFAR10 on VGG5. With inherent sparsity, the sparse SNN now consumes $1.27 \times$ more total energy compared to sparse ANN. Specifically, sparse SNN consumes $1.19\times$ more total computation energy and $1.27\times$ more total memory movement energy compared to sparse ANN. Compared to non-sparse SNN, SATA increases the computation energy efficiency of sparse SNN by $5.58\times$ by utilizing the sparsity. In Fig. \ref{fig:snn_ann_comp}, we visualize the energy comparison results between ANNs and SNNs for both non-sparse and sparse training. We break up the energy results according to Eqn. (7) and (8). We further visualize the layerwise computation energy for the sparse SNN training in Fig. \ref{fig:layerwise_vgg5} and the break up of the total memory energy in Fig. \ref{fig:memory_break}.

We make the following key observations from the comparison results: 
\begin{itemize}
\item We first identify that, in contrast to our impression that SNN is more energy-efficient than ANN, SNN training is more expensive ($1.27\times$ more even with sparsity) than ANN training. Separating the total training energy into computation and memory portions, we observe that though we can utilize the rich sparsity in SNNs to shrink the computation energy gap between SNNs and ANNs ($3.28\times$ to $1.19\times$), the total energy gap ($1.27\times$) is still bounded by the memory energy gap  ($1.27\times$) between two types of networks.
\item With the previous observation, we then identify that the memory communication energy is the bottleneck of the total energy consumption in SNN training. This is due to the expensive cost of accessing to GLBs and DRAMs, which together compose $96.3\%$ of the total memory energy as shown in Fig. \ref{fig:memory_break}.
While memory energy dominates the energy gap between SNNs and ANNs, sparsity hardly optimizes this energy inefficiency. $S$ and $\nabla f$ sparsity can only reduce the memory reads from scratch pads inside PEs but can not optimize the cost of accessing DRAMs and GLBs, which are the most expensive operations in SNN training. 
Moreover, the access to DRAMs needs to be repeated multiple timesteps for reading and writing the necessary data ($S$ and $U$, etc.) for BPTT. In our experiments, due to the small number of timesteps ($T = 8$), the memory access energy for ANNs and SNNs is mainly bounded by the DRAM access energy of filters ($78\%$ of the total memory energy as shown in Fig. \ref{fig:memory_break}), which is the same for both networks. We will show in the later section that larger timesteps will exponentially separate the memory access energy gap between ANN and SNN.
\item We further break up the computation energy into three computation stages to identify the computation energy bottleneck for sparse SNN training. In fact, sparse SNN consumes only $0.26\times$ and $0.44\times$ of sparse ANN's computation energy on the forward and weight update stage. The major bottleneck for SNN's training computation is the backward stage where sparse SNN consumes $2.74\times$ more energy than sparse ANN. During the backward computation, SNNs require the same multi-bits MAC operation as ANNs but the operation needs to be repeated for multiple timesteps. This repetition of MAC operations is the source of computation energy inefficiency in SNN's backward computation.
\item Though the memory energy bottleneck can not be easily fixed with sparsity, the bottleneck for computation energy (namely, backward stage) can be alleviated with sparsity in $\nabla U$. The backward stage of the sparse SNN consumes $0.19\times$  reduced energy than that of the non-sparse SNN. By increasing the $\nabla U$ sparsity, the energy cost of the backward computation stage can be further reduced (refer to energy cost model in  Eqn. (\ref{eq:8})).
\end{itemize}

Fortunately, SNNs not only are highly sparse in spikes but also inherently possess high sparsity in $\nabla U$. We further make the ablation studies on the sparsity of SNNs and their relationship with SNN training energy in the following section.

\subsection{Ablation Study on Sparsity and Training Energy}

\subsubsection{Sparsity and Datasets}

\begin{table}[t]
\centering
    \caption{Accuracy and layerwise sparsity for VGG5. The sparsity shown is the average sparsity per image per timestep. Z denotes the ReLU activation output from ANN.}
\begin{adjustbox}{max width =\linewidth}
    \begin{tabular}{|l|c|c|c|c|c|c|c|}
        \hline
        \textbf{Dataset}&\textbf{Sparsity}& \multicolumn{6}{c|}{\textbf{Layerwise Results (\%)}}\\
        \hline
        &&\textbf{inp} & \textbf{cov1} & \textbf{cov2} & \textbf{cov3} & \textbf{lin4} & \textbf{lin5}\\
        \hline
        \hline
        MNIST &S sparsity&68.83 & 93.03 & 91.98 & 98.06 & 92.91 & -\\
        Acc: 99\% &$\nabla f$ sparsity& - & 22.87 & 38.19 & 55.11 & 32.75 & 46.30\\
        (SNN)&$\nabla U$ sparsity&- & 94.14 & 85.02 & 94.65 & 67.93 & 57.01\\
        \hline
        CIFAR10 &S sparsity&43.45 & 85.83 & 91.57 & 98.37 & 96.82 & -\\
        Acc: 75\% &$\nabla f$ sparsity&- & 39.33 & 69.79 & 80.95 & 62.20 & 37.18\\
        (SNN)&$\nabla U$ sparsity&- & 73.25 & 69.58 & 93.12 & 61.65 & 4.04\\
        \hline
        CIFAR10 &Z sparsity&0.00 & 50.72 & 54.41 & 83.05 & 69.22 & -\\
        Acc: 82\% &$\nabla Z$ sparsity&- & 75.67 & 3.51 & 75.51 & 1.07 & 41.71\\
        (ANN)&&&&&&&\\
        \hline
        CIFAR100 &S sparsity&47.16 & 86.17 & 89.58 & 98.47 & 94.32 & -\\
        Acc: 42\%&$\nabla f$ sparsity& - & 35.20 & 66.65 & 86.66 & 55.58 & 54.19\\
        (SNN)&$\nabla U$ sparsity&- & 70.10 & 65.09 & 95.00 & 54.66 & 10.58\\
        \hline
 	\end{tabular}\label{tab: Sparsity_result}
\end{adjustbox}
\end{table}

\begin{figure}[t]
  \centering
  \includegraphics[width=\linewidth]{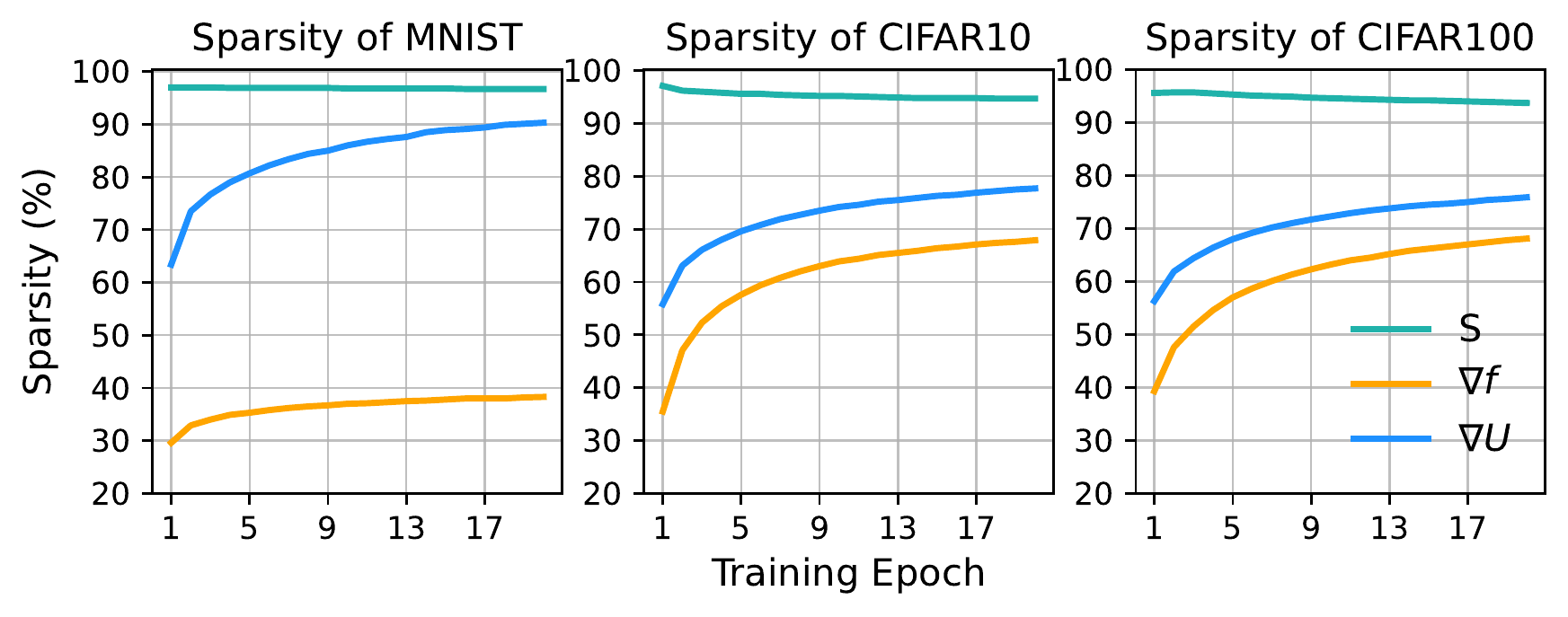}
  \vspace*{-8mm}
  \caption{Sparsity results across datasets.}
  \label{fig:sparsity_dataset}
\end{figure}

We first study the effects of different datasets on SNN's sparsity. We train our VGG5 SNN model across three datasets: MNIST, CIFAR10, and CIFAR100, with the same configurations as in the previous section to generate sparsity results in the first 20 training epochs. For each epoch, each type of sparsity is calculated by averaging across images and timesteps. The results are illustrated in Fig. \ref{fig:sparsity_dataset}. We also provide layerwise sparsity results for three datasets in Table \ref{tab: Sparsity_result}. Several points can be inferred: 
\begin{itemize}
\item First, regardless of the choice of datasets, the spikes ($S$ sparsity) are highly sparse ($>94\%$) throughout the training, which can help SNN save its computation energy during the forward and weight update stages.
\item Second, SNNs also possess a relatively high percentage of $\nabla U$ sparsity (on average $73\%$ on CIFAR10 and $84\%$ on MNIST), which can help SNNs reduce the computation energy for the backward stage.
\item Furthermore, the sparsity of $\nabla f$ and sparsity of $\nabla U$ share similar increasing trends with the increasing number of training epochs. The sparsity-increasing effect is more significant on complex training data (CIFAR100) as compared to the simple one (MNIST).
\end{itemize}

\subsubsection{Sparsity and SNN-unique Hyperparameters}

\begin{figure}[t]

  \centering
  \includegraphics[width=\linewidth]{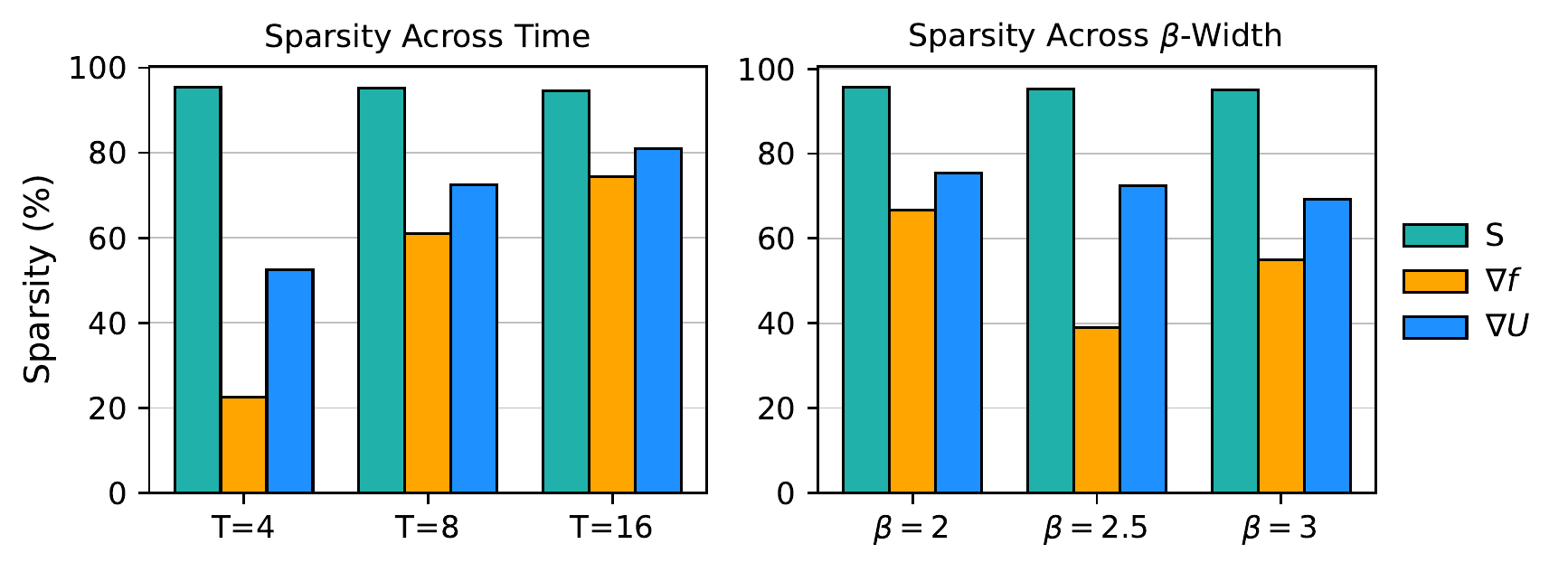}
    \vspace*{-7mm}
  \caption{Sparsity across timesteps and firing function width.}
  \label{fig:sparsity_hyper}
\end{figure}

\begin{figure}[t]
  \centering
  \includegraphics[width=\linewidth]{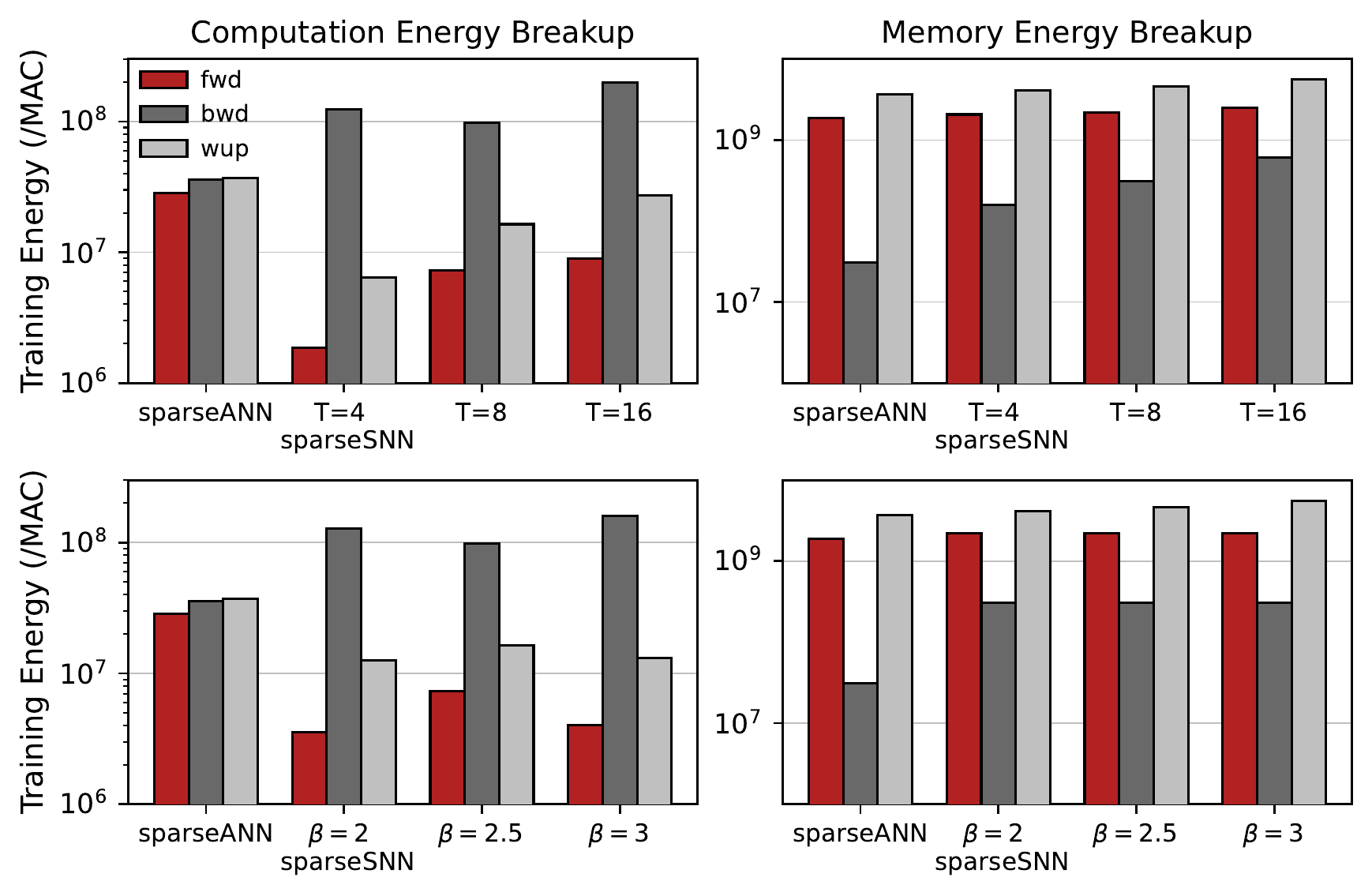}
  \vspace*{-7mm}
  \caption{Energy results across timesteps and firing function width.}
  \label{fig:energy_hyper}
\end{figure}

We further study how the training hyperparameters that are unique to SNNs (namely, timestep $T$ and firing width $\beta$ in Fig. \ref{fig:firing_derivative}) affect the sparsity and the training energy. We train our VGG5 SNN model with different $T$ and $\beta$ to get different sparsity results, as shown in Fig. \ref{fig:sparsity_hyper}. Fig. \ref{fig:energy_hyper} shows the corresponding energy results on sparse ANN and sparse SNN for a different choice of hyperparameters that result in different levels of sparsity.

As shown in Fig. \ref{fig:energy_hyper}, changing firing width has almost no effect on the SNN training energy. Also, naively adjusting $T$ does not result in a proportional change in computational energy. For example, while reducing $T$ reduces the number of repeated computation operations, it also reduces the sparsity of $\nabla f$ and $\nabla U$, and thus cancels out the saved energy from reduced computation operations. As we discussed in section 7.2, the memory communication energy is bounded by the movement of filters on our VGG5 example. Thus, we find that only the backward memory cost (which does not involve movements of filters) is proportional to the number of timesteps. We will have further discussions on the effects of the timestep in the next section.

\subsubsection{Sparsity and Network Depth}
\begin{figure}[t]

  \centering
  \includegraphics[width=\linewidth]{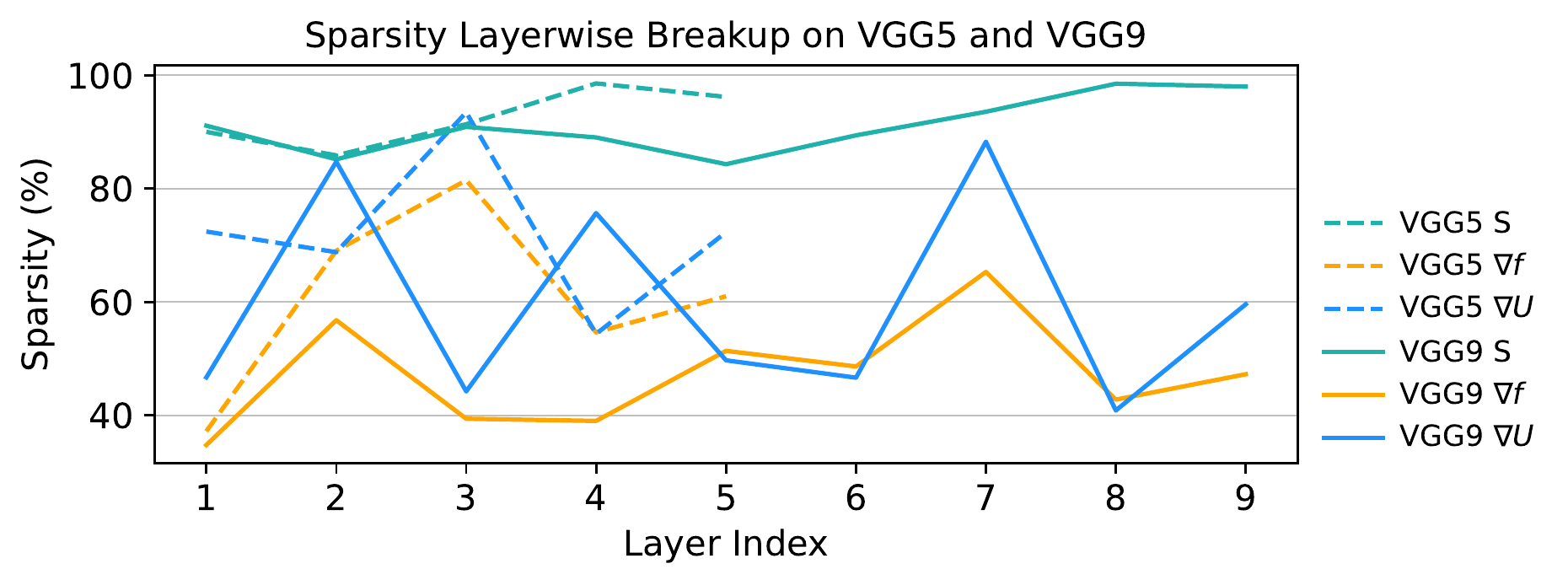}
  \vspace*{-8mm}
  \caption{Layerwise sparsity comparison between VGG5 and VGG9.}
  \label{fig:vgg9_sparsity}
\end{figure}

\begin{figure}[t]
  \centering
    \includegraphics[width=0.8\linewidth]{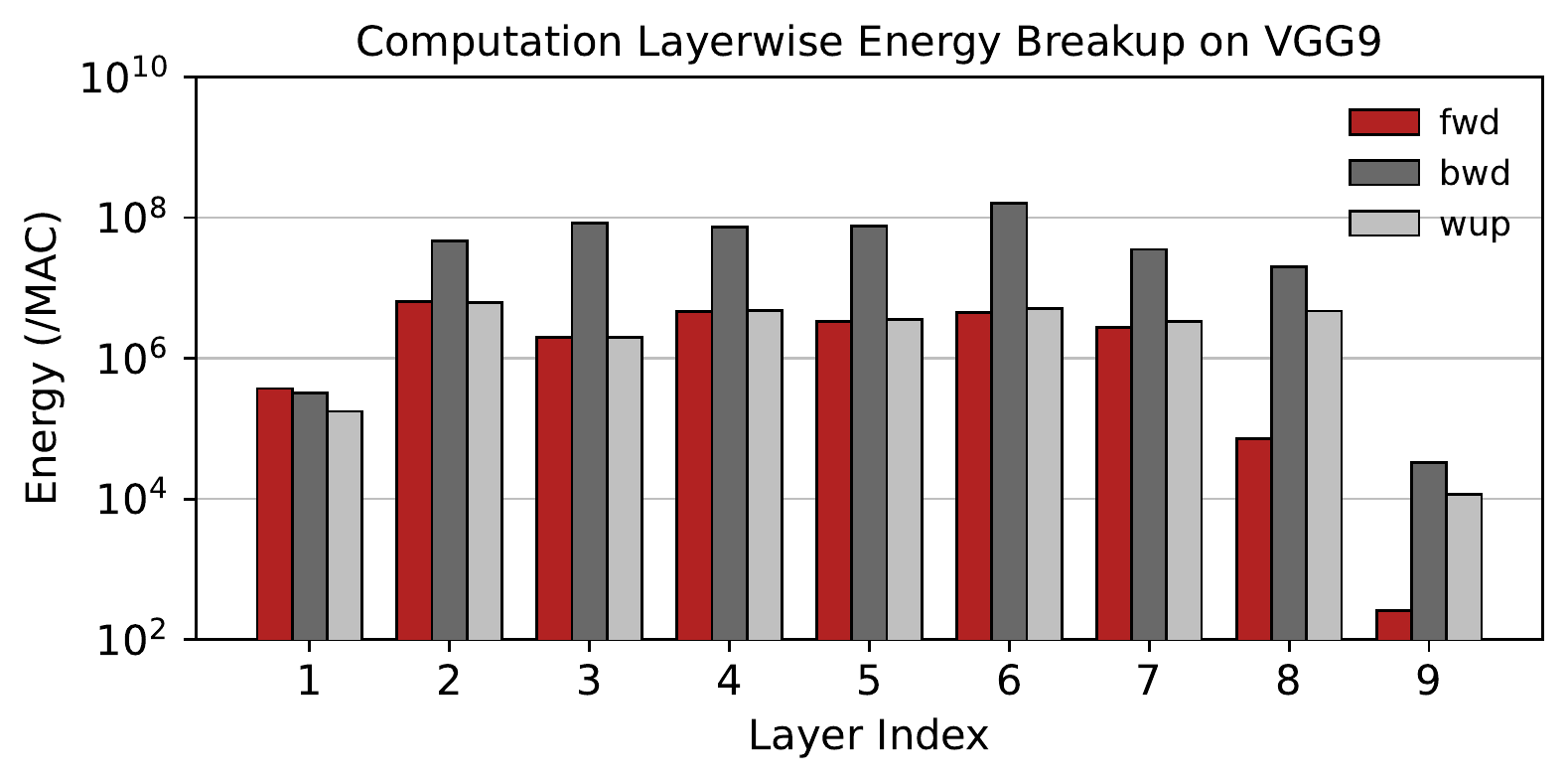}
    \vspace*{-3mm}
  \caption{Layerwise computation energy results on VGG9.}
  \label{fig:vgg9_energy}
\end{figure}

Finally, we study the effects of network depth and sparsity. We further train a VGG9 network with the same training configurations as our previous VGG5 model on CIFAR10 and get the average layerwise sparsity results, as shown in Fig. \ref{fig:vgg9_sparsity}. We observe that, while $S$ sparsity gets more sparse in the deeper layers, the changing trend and average sparsity across layers are roughly the same for both networks. For $\nabla U$ sparsity, both networks also share a similar changing trend across layers. On average, VGG9 experiences less $\nabla U$ sparsity ($\sim60\%$) across layers compared to VGG5 ($\sim70\%$). We generate the layerwise computation energy with our energy estimation model and visualize the results in Fig. \ref{fig:vgg9_energy} for VGG9.

\begin{figure}[t]
  \centering
\includegraphics[width=0.85\linewidth]{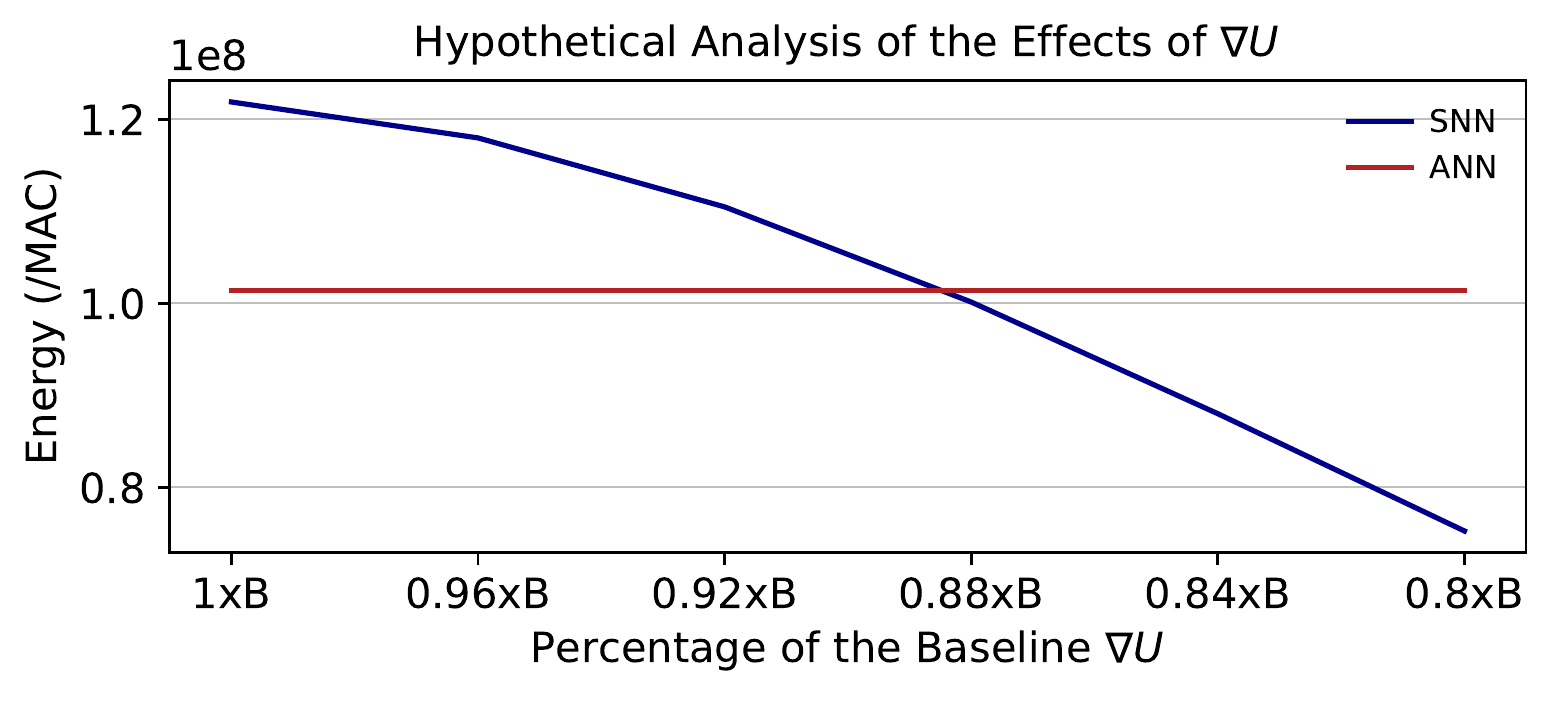}
\vspace*{-3mm}
  \caption{Hypothetical analysis on how $\nabla U$ affects energy efficiency. $B$ denotes the layerwise density i.e. $B=1- sp_{\nabla U}$. $sp_{\nabla U}$ denotes the layerwise sparsity due to $\nabla U$.}
  \label{fig:du_hypothetical}
\end{figure}

\begin{figure}[t]
  \centering
\includegraphics[width=0.85\linewidth]{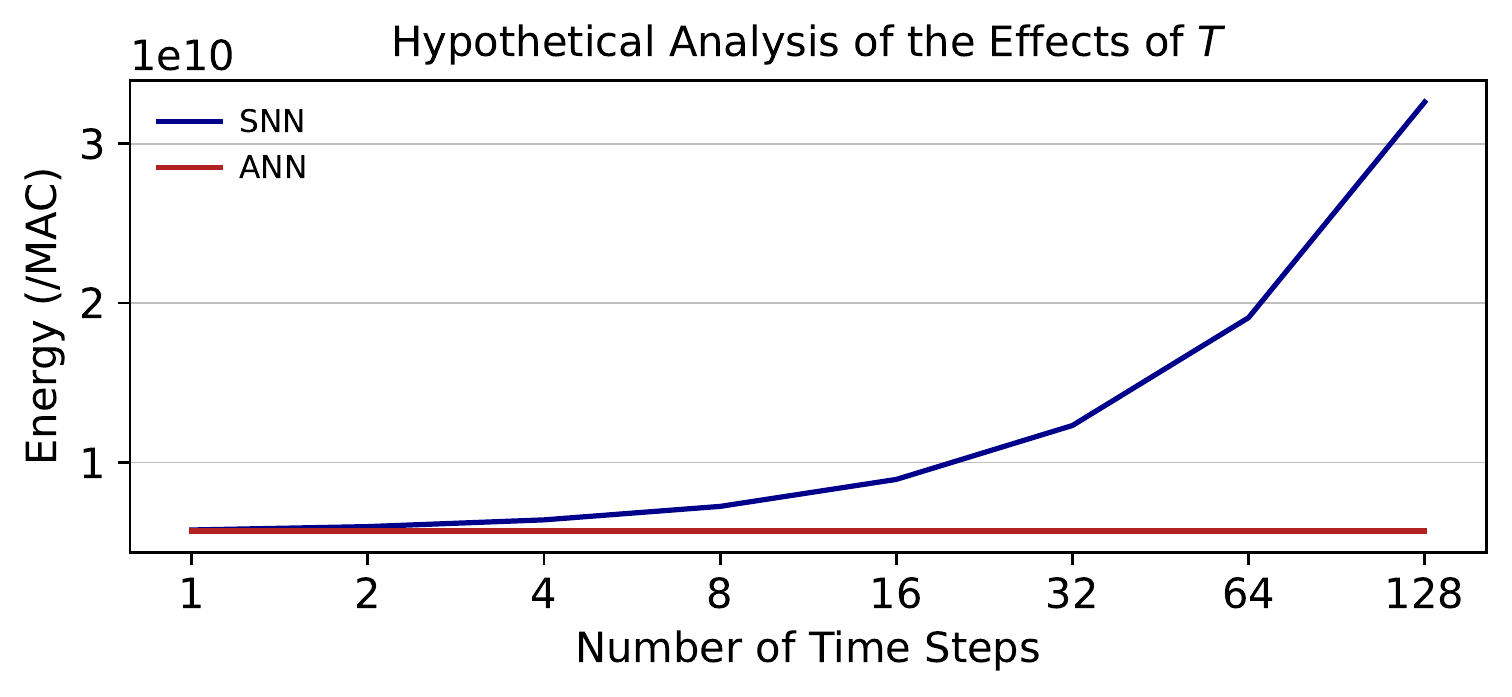}
\vspace*{-3mm}
  \caption{Hypothetical analysis of total energy cost on total timesteps $T$.}
  \label{fig:t_hypothetical}
\end{figure}

\subsection{Discussion}
\subsubsection{SNN training algorithm}
In this section, we further discuss some possible future directions for SNN algorithm design to make SNN training energy-efficient. One direction would be to optimize the total computation energy. As discussed in Section 7.2, the bottleneck for SNN training computation energy is backward computation. This bottleneck can be alleviated by introducing more $\nabla U$ sparsity during the training. While simply adjusting the training parameters can not effectively increase the $\nabla U$ sparsity, we provide a hypothetical analysis to show the tradeoff between the total computation energy and the $\nabla U$ sparsity in Fig. \ref{fig:du_hypothetical}. We use the sparse ANN training energy as in section 7.2 and fix it. We take the $\nabla U$ layerwise sparsity of CIFAR10 on VGG5 SNN in Table \ref{tab: Sparsity_result} as our baseline sparsity and gradually scale it up.% ($1-B$). Then, we gradually introduce more $\nabla U$ sparsity by proportionately scaling down the density ($B$). All other conditions are fixed. 
We observe that by increasing the $\nabla U$ sparsity, SNN training will have less total computation energy overhead compared to ANN training. At 88\% of baseline, the SNN breaks even with ANN. 

To optimize the total training energy of SNN, a large number of timesteps should be avoided. We make a similar hypothetical analysis as above on the relation between total timesteps $T$ and training energy of SNNs in Fig. \ref{fig:t_hypothetical}. We find that SNN's total training energy exponentially increases with the number of timesteps. This is because we need to repetitively access DRAMs for $T$ times for getting membrane potential ($U$) and spike ($S$) for BPTT. This expensive memory operation will dominate the total energy when $T$ gets large. Apart from the energy dominance, the training time of SNN will also increase as timesteps increase. Table \ref{tb:train_time} shows how the training latency gap between SNNs and ANNs gets bigger when the timesteps increase.

\begin{table}[h]
\centering
    \caption{Latency comparison of one training epoch between SNNs and ANNs with varying timesteps over VGG5 on NVIDIA V100 GPU.}
    \begin{adjustbox}{max width =\linewidth}
	\begin{tabular}{|l|c|c|c|}
        \hline
        \textbf{Network}&\textbf{Latency of ANNs}&\textbf{Latency of SNNs}&\textbf{Latency Gap}\\
        \hline
        \hline
        VGG5&12.28 s&83.29 s ($T=4$)&$6.78\times$\\
        \hline
        VGG5&12.28 s&180.12 s ($T=8$)&$14.67\times$\\
        \hline
        VGG5&12.28 s&409.92 s ($T=16$)&$33.38\times$\\
        \hline
 	\end{tabular}\label{tb:train_time}
 	\end{adjustbox}
\end{table}

Moreover, as we have shown in Fig. \ref{fig:memory_break}, energy for DRAM data movement of the weights becomes the bottleneck of the SNN training. One possible future direction is to train the SNNs with a sparsity constraint. The other possibility is to compress part of or even the whole model through methods like \cite{chen2019eyerissv2, kim2022lottery,deng2021gospa} and store the model on-chip as in \cite{truenorth}.

\subsubsection{SNN training accelerator $\&$ comparison with prior work} 
\begin{figure}[t]
  \centering
\includegraphics[width=0.55\linewidth]{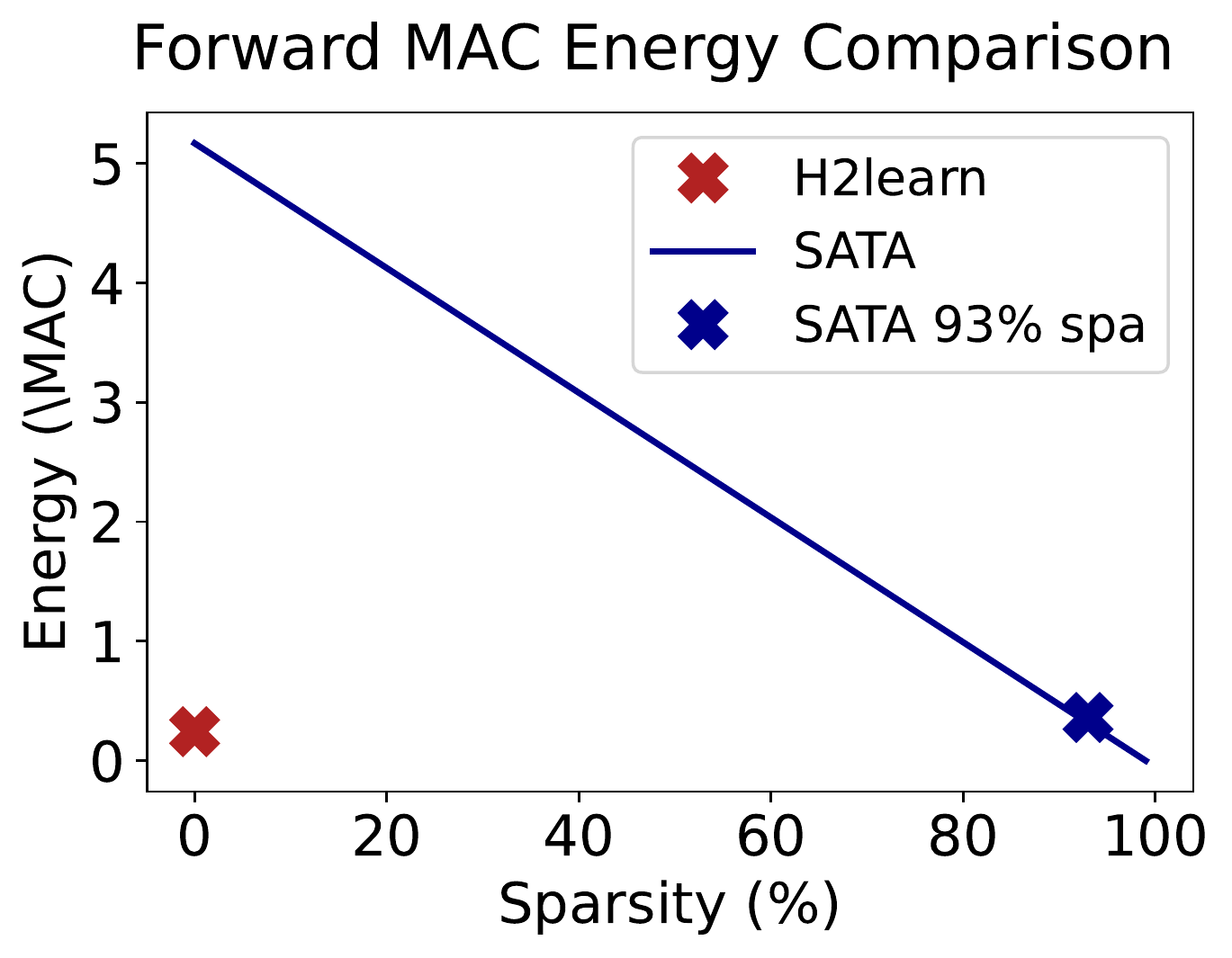}
  \vspace*{-3mm}
  \caption{Energy comparison between SATA and H2Learn for a forward convolution workload ($3\times 3$ kernel).}
  \label{fig:h2learn_sata}
\end{figure}

In this section, we discuss some considerations for the future design of SNN training accelerators based on the findings from this paper. From our energy comparison results, we find SNNs are less energy efficient than ANNs in a gradient-based training setup. SATA being a general purpose architecture targeted to perform fast energy estimation and comparison between different SNN structures, we do not pay much effort to the architectural level optimization for BPTT-based SNN training, except for the sparsity-aware PEs and PGUs.
One future direction for the SNN training accelerator design would be optimizing the time-repetitive data movement for the BPTT-based method. For instance, the SNN-dedicated design proposed in H2Learn \cite{liang2021h2learn} indeed unveils some potential ways to alleviate the memory movement bottleneck for SNNs.
We implement the LUT-based PE from the Forward Engine in H2Learn\cite{liang2021h2learn} with 65nm CMOS technology and use the same synthesis method as SATA. We compare the energy difference between SATA's PE and LUT PE on performing a convolution using a $3\times 3$ kernel for one timestep. The energy difference is shown in Fig.\ref{fig:h2learn_sata}. Due to the LUT-based convolution that H2Learn utilizes, the energy result for the convolution in SNN's forward propagation does not suffer from the time-repetitive memory reading from scratchpads inside PEs. Also, the LUT-based convolution is sparsity-independent. Thus, SATA's general purpose PE consumes approximately $21.2 \times$ more energy on a $3\times 3$ convolution workload without considering sparsity. When considering the sparsity, SATA can only get the same energy efficiency as H2Learn with $93\%$ sparsity (not possible for a $3 \times3$ kernel that delivers information).
% When we consider multiple timesteps, SATA will suffer more from the sparsity-dependent effect.
Note, the above comparison is approximate, for example, the energy overheads of pre-calculating and loading the elements for LUTs are not considered. Indeed, considering those overheads would make the energy estimation and comparison between the training of different SNN structures complex. That's also one major motivation for having SATA, a general purpose architecture design for simple SNN training energy estimation and comparison.

\section{Conclusions}
We propose SATA, a sparsity-aware BPTT-based training accelerator for SNNs. The simple and highly re-configurable systolic-based design of SATA makes it easy to perform a training energy analysis on different SNN topologies. We further propose an energy estimation model based on SATA for energy estimation. Compared with not utilizing sparsity, sparsity-aware SATA increases its computation energy efficiency by $5.58 \times$. The results also show that when running on Eyeriss-like systolic-based architecture, SNN training requires more energy compared to ANNs with and without considering sparsity. We make several observations and show how energy-efficiency trade-off with respect to different SNN-specific training parameters. Our results and estimation tool will hopefully guide future SNN algorithm works to design more energy-efficient and sparsity-aware training mechanisms, as well as future SNN training accelerator works to improve their architecture design to be more energy-efficient.

%By using the energy model, we make a training energy comparison between SNNs and ANNs.

\section*{Acknowledgments}
The research was funded in part by C-BRIC, one of six centers in JUMP, a Semiconductor Research Corporation (SRC) program sponsored by DARPA, the National Science Foundation (Grant$\#$1947826), DARPA AI Exploration (ShELL), and, Technology Innovation Institute (Abu Dhabi).

%TII DARFAA-AI

\bibliographystyle{IEEEtran}
\bibliography{main}

\begin{IEEEbiography}[{\includegraphics[width=1in,height=1.25in,clip,keepaspectratio]{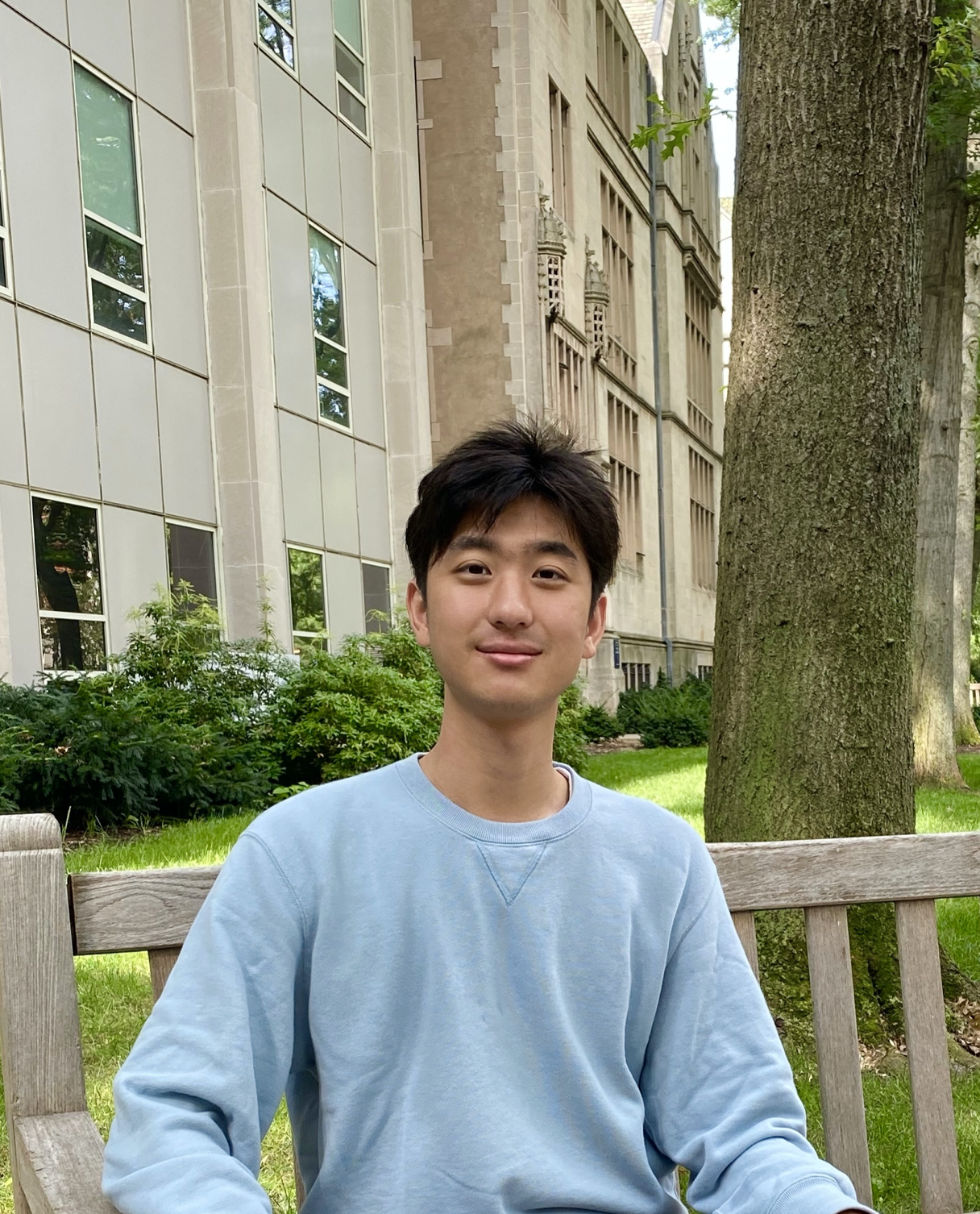}}]{Ruokai Yin} is a Ph.D. student in the Department of Electrical Engineering at Yale University, advised by Prof. Priyadarshini Panda. His research interests lie in designing high-performance computer architectures for neural networks. Prior to joining Yale, he received his BS-Electrical Engineering degree from the University of Wisconsin-Madison, where he worked with Prof. Joshua San Miguel on computer architectures for stochastic computing.
\end{IEEEbiography}

\begin{IEEEbiography}[{\includegraphics[width=1in,height=1.25in,clip,keepaspectratio]{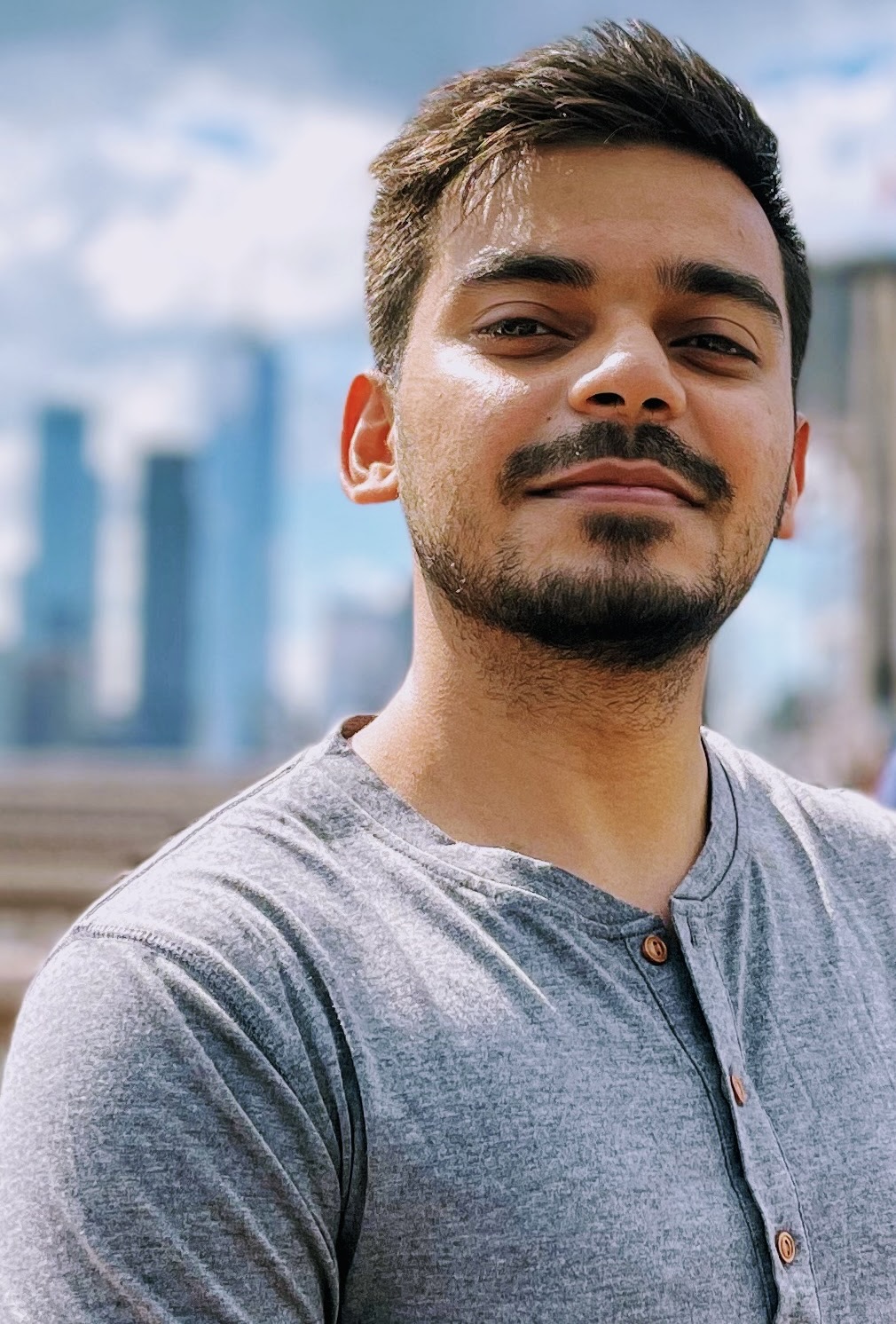}}]{Abhishek Moitra} received his B.E. degree in Electrical Engineering from Birla Institute of Technology and Science Goa, India in 2019. Currently, he is pursuing his Ph.D. in the Intelligent Computing Lab at Yale. Previously, he worked as a research assistant at the Indian Institute of Science, Bangalore where he worked on designing FPGA-based hardware accelerators for Signal and Image processing applications. His research interests involve hardware-algorithm co-design with CMOS and emerging devices for efficient and robust Deep Learning. 
\end{IEEEbiography}

\begin{IEEEbiography}[{\includegraphics[width=1in,height=1.25in,clip,keepaspectratio]{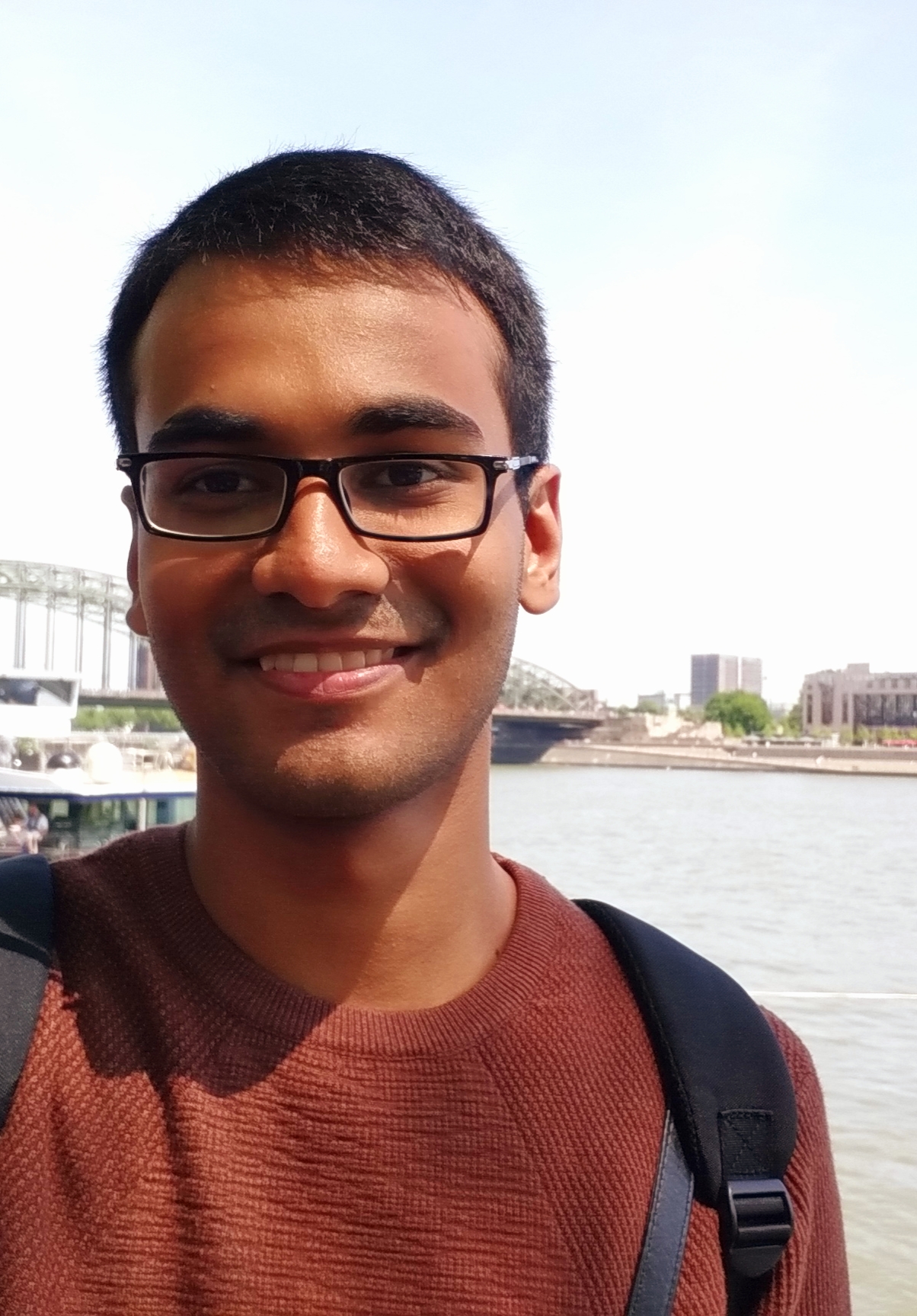}}]{Abhiroop Bhattacharjee} received B.E.\ in Electrical and Electronics from Birla Institute of Technology and Science Pilani, India, in 2020. He joined Yale University, USA, in 2020 as a Ph.D. student in the Electrical Engineering department. Prior to joining Yale University, he worked as a guest researcher in the Chair for Processor Design, TU Dresden, Germany, in 2020, and as a research intern in the Institute of Materials in Electrical Engineering-I, RWTH Aachen University, Germany, in 2019. His research interests lie in the areas of adversarial security and process in-memory architectures for neuromorphic circuits.
\end{IEEEbiography}

\begin{IEEEbiography}[{\includegraphics[width=1in,height=1.23in,clip,keepaspectratio]{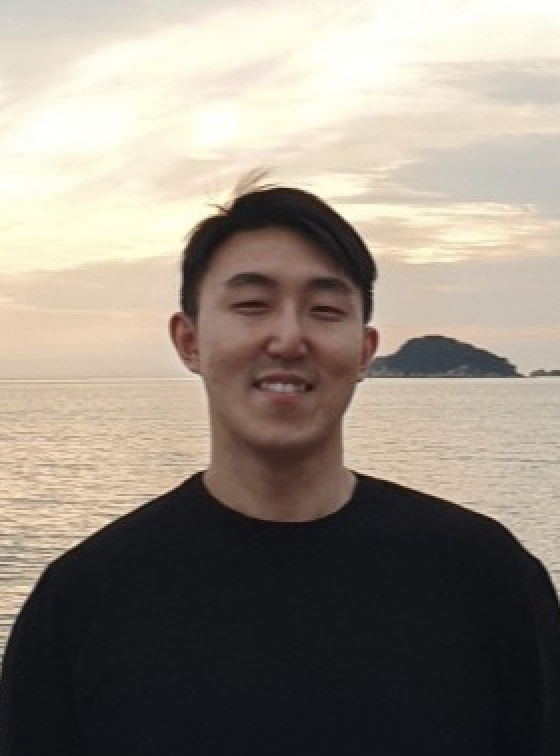}}]
{Youngeun Kim}
is currently working toward a Ph.D. degree in Electrical Engineering at Yale University, New Haven, CT, USA.
Prior to joining Yale,  he worked as a full-time student intern at T-Brain, AI Center, SK telecom, South Korea. 
He received his B.S. degree in Electronic Engineering from Sogang University, South Korea, in 2018 and M.S. degree in Electrical Engineering from Korea Advanced Institute of Science and Technology (KAIST), in 2020. 
His research interests include neuromorphic  computing, computer vision, and deep learning.
\end{IEEEbiography}

\begin{IEEEbiography}[{\includegraphics[width=1in,height=1.25in,clip,keepaspectratio]{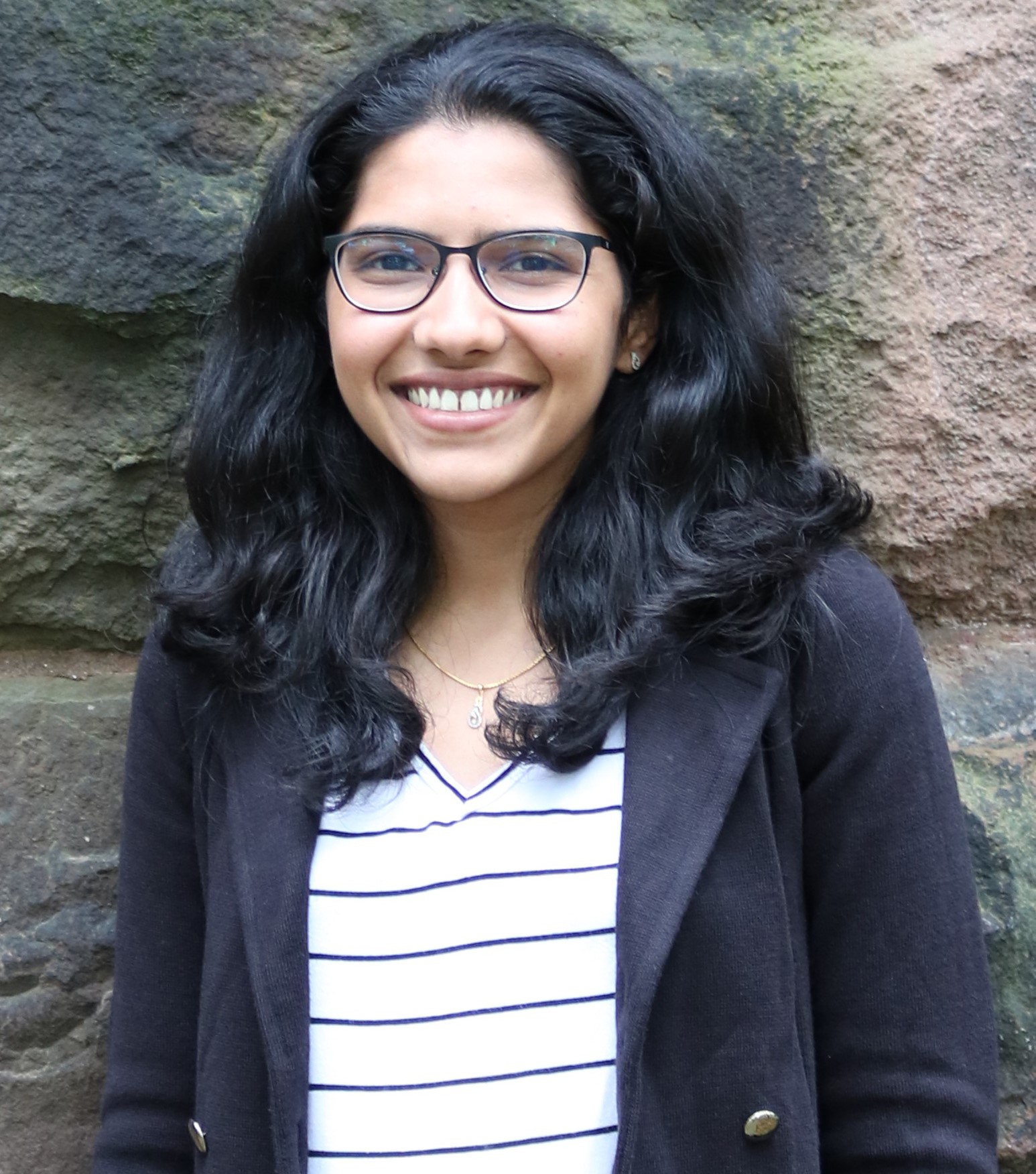}}]{Priyadarshini Panda} is an assistant professor in the electrical engineering department at Yale University, USA. She received her B.E. degree in Electrical \& Electronics and Master's degree in Physics from BITS, Pilani, India in 2013 and her PhD in Electrical \& Computer Engineering from Purdue University, USA in 2019. She was the recipient of outstanding student award in Physics in 2013. From 2013-14, she worked at Intel, India as a design engineer and Nvidia, India as an intern. In 2017, she interned in Intel Labs, Oregon, USA where she developed large scale spiking neural network algorithms for benchmarking the Loihi chip. She is the recipient of the 2019 Amazon Research Award, 2022 Google Scholar Research Award, and 2022 DARPA Riser Award. She has published more than 60 publications in well-recognized venues including, Nature, Nature Communications, and IEEE among others. Her research interests include- neuromorphic computing, energy efficient deep learning, adversarial robustness, and hardware-centric design of robust neural systems.
\end{IEEEbiography}

% \newpage
 
\vspace{11pt}

\vspace{11pt}

\vfill

\end{document}